\documentclass[acmsmall]{acmart}
\AtBeginDocument{%
  \providecommand\BibTeX{{%
    \normalfont B\kern-0.5em{\scshape i\kern-0.25em b}\kern-0.8em\TeX}}}

\setcopyright{acmcopyright}
\copyrightyear{2023}

\usepackage{tikz}
\usepackage{multirow} 
\usepackage{graphicx} 
\usepackage{float}
\usepackage{subfigure}
\usepackage{enumitem}
\usepackage{titlesec}

\usetikzlibrary{trees,positioning,shapes,shadows,arrows}
\tikzset{
  basic/.style  = {draw, text width=2cm, drop shadow, font=\sffamily, rectangle},
  root/.style   = {basic, rounded corners=2pt, thin, align=center, fill=white},
  level-2/.style = {basic, rounded corners=6pt, thin,align=center, fill=white, text width=3cm},
  level-3/.style = {basic, thin, align=center, fill=white, text width=1.8cm}
}

\tikzset{
  treenode/.style = {shape=rectangle, rounded corners, draw, align=center, top color=white, drop shadow,xshift=0cm,minimum width=2cm},
  root2/.style     = {treenode, font=\Large, bottom color=white},
  env/.style      = {treenode, font=\ttfamily\normalsize},
  dummy/.style    = {circle,draw}
}
\begin{document}

\title{Graph Neural Network for spatiotemporal data: methods and applications}

\author{Yun Li}
\email{yli230@emory.edu}
\authornote {Authors contributed equally to this research}
\author{Dazhou Yu}
\authornotemark[1]
\email{dazhou.yu@emory.edu}
\author{Zhenke Liu}
\authornotemark[1]
\email{zhenke.liu@emory.edu}
\author{Minxing Zhang}
\email{minxing.zhang@emory.edu}
\author{Xiaoyun Gong}
\email{kristina.gong@emory.edu}
\author{Liang Zhao}
\email{liang.zhao@emory.edu}
\affiliation{%
  \institution{Emory University}
  \city{Atlanta}
  \state{Georgia}
  \country{USA}
}

\renewcommand{\shortauthors}{Li and Yu, et al.}

\begin{abstract}
In the era of big data, there has been a surge in the availability of data containing rich spatial and temporal information, offering valuable insights into dynamic systems and processes for applications such as weather forecasting, natural disaster management, intelligent transport systems, and precision agriculture. Graph neural networks (GNNs) have emerged as a powerful tool for modeling and understanding data with dependencies to each other such as spatial and temporal dependencies. There is a large amount of existing work that focuses on addressing the complex spatial and temporal dependencies in spatiotemporal data using GNNs. However, the strong interdisciplinary nature of spatiotemporal data has created numerous GNNs variants specifically designed for distinct application domains. Although the techniques are generally applicable across various domains, cross-referencing these methods remains essential yet challenging due to the absence of a comprehensive literature review on GNNs for spatiotemporal data. This article aims to provide a systematic and comprehensive overview of the technologies and applications of GNNs in the spatiotemporal domain. First, the ways of constructing graphs from spatiotemporal data are summarized to help domain experts understand how to generate graphs from various types of spatiotemporal data. Then, a systematic categorization and summary of existing spatiotemporal GNNs are presented to enable domain experts to identify suitable techniques and to support model developers in advancing their research. Moreover, a comprehensive overview of significant applications in the spatiotemporal domain is offered to introduce a broader range of applications to model developers and domain experts, assisting them in exploring potential research topics and enhancing the impact of their work. Finally, open challenges and future directions are discussed.

\end{abstract}




\maketitle

\section{Introduction}
\noindent An unprecedented amount of data with rich spatial and temporal information have become available due to the advanced development of IoT sensors, satellites, and relevant sensing technologies. Big spatiotemporal data provide valuable insights into dynamic systems and processes and are useful for various applications such as environmental monitoring, natural disaster management, weather nowcasting and forecasting, intelligent transport systems, and precision agriculture \cite{yang2019big}. Spatiotemporal mining, which involves detecting and predicting patterns and trends in data collected over space and time in unsupervised and supervised manners, plays a critical role in system understanding and knowledge discovery to support decision-making in fields such as natural and social science \cite{shekhar2015spatiotemporal, atluri2018spatio, shekhar2015spatiotemporal}. However, spatiotemporal data poses grant challenges to traditional machine learning models as this kind of data is inherently correlated in both space and time which makes it difficult to efficiently model the underlying dynamics and patterns with traditional machine learning models designed for independent and identically distributed (i.i.d.) data. With the rapid development of deep learning, neural networks have been widely adopted to support spatiotemporal data mining due to their capability of learning latent features, recognizing patterns, and solving classification and prediction problems\cite{wang2020deep, ghaderi2017deep}. For example, deep neural networks and their variants are utilized for modeling the non-linear relationships among features of spatial points \cite{le2020spatiotemporal}, Convolutional Neural Network (CNN) was adopted to extract high-level representations from spatial imagery for downstream tasks \cite{yao2019review}, Recurrent Neural Networks (RNN) has been implemented to capture the short-term and long-term linkages in a time series dataset \cite{fang2021survey}. These AI technologies are powerful to handle predefined features and regular data relations in tabular, image, and sequential data. However, spatiotemporal data requires the power to explore the mutual dependencies between locations and spatial correlations that can be irregular, heterogeneous, and non-stationary, which hence inherently require much more powerful and expressive techniques.

In recent years, GNN has been proven to be a powerful tool for modeling and understanding spatiotemporal data and has great potential to improve our ability to make predictions and forecasts based on spatiotemporal data since it incorporates graph-based representations into the neural network \cite{wu2020comprehensive}. GNN is a class of neural networks specifically designed to operate on graph-structured data, such as a social network, a transportation network, a chemical compound. It takes into account the inherent spatial dependencies present in graph-based data structures and is particularly useful for capturing the complex relationships and dependencies among the nodes in a graph, such as the interactions between different spatial regions. GNN is quite useful for tasks such as node classification, link prediction, and graph classification such as place of interest recommendation, teleconnection discovery in large-scale spatiotemporal datasets, and community detection in a graph \cite{xu2018powerful}. A model integrating GNN with RNNs such as Long short-term memory (LSTM) and Gated recurrent unit (GRU) could capture spatial dependency and temporal dependency simultaneously, thus it is able to perform a wide range of spatiotemporal mining tasks.

The field of GNN is currently experiencing tremendous growth in spatiotemporal applications \cite{zhou2021ast, li2021multiscale, kapoor2020examining,sun2021adaptive}, driven by the combination of recent advancements in data collection techniques (e.g., in-situ and remote sensors, social sensors), the development of new GNN architectures (e.g. Graph Convolutional Network, Graph Attention Network, Graph Isomorphism Network) and the availability of high-performance computing platforms. Despite the promise of GNNs in spatiotemporal data mining,
applying GNNs to the domain requires the invention and integration of related techniques to address open challenges caused by the unique characteristics of spatiotemporal data and applications including: 
(1) {\bfseries Graph construction from spatiotemporal inputs}. Building a graph involves finding an appropriate representation of nodes from input and determining the relationships between nodes. However, not all spatiotemporal data have an explicit graph structure, which can create challenges in constructing a graph from non-graph structure data, so how can we incorporate prior knowledge in a domain to aid in establishing a graph structure? Even for data with explicit graph structures such as road networks, how can we select node and edge features to determine information flow between nodes? It is also challenging to determine whether to build a static graph or a dynamic graph and how to build an adaptive dynamic graph. (2) {\bfseries Handling of spatial and temporal dependencies simultaneously}. GNN is well suited for capturing spatial dependencies, but it often neglects the temporal dimension. However, temporal dependencies play a critical role in spatiotemporal mining tasks, specifically for spatiotemporal prediction. For example, the intensity of a tropical cyclone at a specific time step is highly related to its intensity in the previous time steps \cite{li2017leveraging}. So, how should we incorporate temporal dependencies into GNNs to model spatiotemporal data? (3) {\bfseries Challenges in the big spatiotemporal data}. Traditional GNN models are often not able to handle large-scale data. For example, when representing a global-scale raster dataset as a graph in which each grid cell is treated as a node, the number of nodes in the graph can be very large, making it difficult to train a GNN model. So how to design more efficient and scalable GNNs to handle big spatiotemporal data?

In recent years, a significant amount of research has been dedicated to the technology development and spatiotemporal applications of GNNs in order to address the aforementioned challenges. There has been a growing trend in the number of studies that propose and apply  GNN models to support spatiotemporal mining in various domains, despite the fact that GNN is still a relatively new field. Many existing GNN models have been designed for a specific domain, but these models can often be transferred to address problems in other domains. However, it can be difficult to cross-reference GNN and its variants across different domains without a systematic survey of the various GNN techniques that have been proposed to support spatiotemporal mining. This also prevents many researchers in this field from a whole picture of the existing techniques and limitations, open problems and challenges, and potential future research directions.

In light of the challenges faced when applying GNNs to spatiotemporal applications, this survey aims to provide a thorough and systematic review of the  state of the art for GNNs in spatiotemporal mining. The main objectives of this article include:

\begin{itemize}[fullwidth]
\item {\textit{\bfseries A systematic taxonomy of a generic framework for GNN models in spatiotemporal mining}: spatiotemporal mining tasks are categorized according to their spatial inputs, temporal dimensions, and corresponding techniques to create the taxonomy of a general framework. The relationships and distinctions among various categories are discussed, including introductions to the models used within each subcategory. The taxonomy could assist researchers in the field to identify the most appropriate techniques for their specific problem contexts.} 
\item {\textit{\bfseries A comprehensive summarization of key spatiotemporal application areas}: A comprehensive list of spatiotemporal applications is provided, focusing on the practical significance and graph formulation for spatiotemporal mining tasks in each domain or sub-domain. This allows domain scientists to easily track the proposed GNN models in a specific domain and can also assist data scientists and model developers in identifying additional areas of application and relevant datasets for evaluating and improving their models.}
\item {\textit{\bfseries Summarization of benchmarks, evaluation metrics, and procedures}:} We summarize and categorize the existing evaluation procedures and metrics, the benchmark datasets, and the corresponding results of GNNs for spatiotemporal prediction tasks. Benchmarks for GNNs in spatiotemporal data have also been comprehensively surveyed in different application fields.

\item {\textit{\bfseries An enlightening discussion of the development of GNN in spatiotemporal domain and future trends}: Through a comprehensive review of existing GNN techniques and applications in spatiotemporal domain, the article provides an understanding of the current research frontiers in the field, highlights challenges and potential issues, and explores potential future directions for GNN research that can aid in spatiotemporal data mining.
}
\end{itemize}

\subsection{Related Surveys}
\noindent This section provides a summary of previous surveys on GNN methods and applications, divided into three groups including (1) GNN methods, (2) GNN applications, and (3)  spatiotemporal data mining. GNN is a rapidly growing field and several models such as GCN, GAT, and GIN have been proposed for graph data processing. There have been a number of surveys evaluating and comparing these GNNs and their variants to provide a comprehensive overview of GNNs. For example, \cite{wu2020comprehensive} presented a survey of GNN architectures and their variants. \cite{oloulade2021graph} presented a survey of graph neural architectures. \cite{thomas2022graph} provided a review of GNNs for different graph types. Several researchers summarized comprehensive reviews of GNN algorithms and applications \cite{wu2020comprehensive, gupta2021graph, liang2022survey}.
GNNs have been extensively explored for machine learning tasks such as node classification, graph classification, and link prediction. These tasks are typically motivated by specific problems, such as recommendation systems. There have been a number of surveys of GNN applications handling a specific task. For example, \cite{kim2022graph} provided a survey of graph anomaly detection with GNNs, \cite{chen2021graph} focused on GNN-based fault diagnosis. Other surveys have focused on specific tasks such as recommender systems \cite{wu2022graph, sharma2022survey, gao2022graph}, graph transformers in computer vision \cite{chen2022survey}, text representation and classification \cite{pham2022deep}, sentiment analysis \cite{luo2021survey}, knowledge graphs \cite{ye2022comprehensive}. Meanwhile, with the rapid development of GNNs, they have been widely employed in various domains. There are a number of surveys focusing on GNN for specific domains such as traffic forecasting \cite{bui2022spatial, jiang2022graph}, finance \cite{wang2021review}, bioinformation \cite{zhu2022survey, li2021graph}, network management and orchestration \cite{tam2022graph}, IoT \cite{dong2022graph}, urban computing \cite{jin2023spatio}. With the swift advancement of sensor technology, data containing rich spatiotemporal information has become widespread across various domains. Many spatiotemporal mining methods have been developed to effectively process spatial and temporal information, revealing valuable insights for decision-making.  Several surveys have been conducted to compare STDM techniques and provide a comprehensive overview of handling spatiotemporal data with advanced data mining and machine learning techniques. For example, \cite{ansari2020spatiotemporal} focused on spatiotemporal clustering techniques, \cite{mazimpaka2016trajectory} summarized trajectory data mining, and \cite{atluri2018spatio} discussed different types of spatiotemporal data and classified spatiotemporal data mining into six major categories. With the development of deep learning techniques, some surveys have focused on deep learning for spatiotemporal data, such as \cite{gao2022generative} introduced how generative adversarial networks have been adopted for spatiotemporal data, and \cite{wang2020deep} provided a comprehensive review of recent progress in applying deep learning techniques for STDM. However, despite the potential and rapid growth of GNN models and spatiotemporal data, the use of GNNs in handling spatiotemporal data still lacks a comprehensive and systematic literature survey covering all its aspects, including relevant techniques, applications, and open problems. However, despite its potential and rapid growth, the use of GNNs in handling spatiotemporal data still lacks a comprehensive and systematic literature survey covering all its aspects, including relevant techniques, applications, and open problems.

\subsection{Outlines}
\noindent  The remainder of this article is organized as follows. Section 2 presents generic graph construction for spatiotemporal data mining. Section 3 introduces a taxonomy and comprehensive descriptions of how GNN techniques support spatiotemporal data mining. Section 4 then summarizes and presents the various applications of GNN using spatiotemporal data, after which Section 5 lists the open challenges and potential future research directions. Section 6 concludes this survey. 

\section{Graph Construction from spatiotemporal data for GNNs}
\noindent  GNN is a specialized type of neural network that is specifically designed to operate on graph-structured data. In spatiotemporal data mining, constructing a suitable graph representation from spatiotemporal data is an essential step in leveraging GNNs to perform data mining tasks. This section delves into graph construction, in an effort to aid researchers in gaining a comprehensive understanding of the methods and techniques for graph representations of spatiotemporal data.

\subsection{Spatiotemporal Data}
\noindent  Vector and raster data are two common ways depicting spatial information in space \cite{anselin2009geoda}. Vector data is a traditional method of cartographic representation and is made up of points, lines, or polygons. Each point in vector data is represented by its spatial coordinates, while lines and polygons are formed by connecting these points in a specific order to represent a discrete object with a clear boundary such as roads, building footprints, and land parcels. On the other hand, raster data represent space by dividing it into a regular grid of cells, with each cell associated with a set of spatial coordinates and one or more values that represent characteristics of the cell, such as elevation, temperature, or land cover type. Raster data are useful for representing data in continuous space. They are commonly used in fields such as remote sensing and geology.
Spatiotemporal data refers to data that include both spatial and temporal information and can be classified into event, trajectory, point reference, raster, and video data\cite{wang2020deep}. It can be represented in vector or raster format at multiple time stamps. For instance, trajectory data can be represented as a sequence of points data. Some spatiotemporal data can be naturally represented as graphs, such as road networks or river networks. However, some spatiotemporal data do not have an explicit graph structure and require preprocessing to identify hidden linkages for graph construction.

\subsection{Graph Data}
\noindent  A graph is a data structure that consists of a set of nodes and a set of edges connecting them. Formally, we define a graph as $G=(V,E)$, where V is the set of nodes and E is a set of edges. Each edge in E connects two nodes in V. The edges can be directed or undirected, depending on whether the relationships they represent are one-way or bidirectional. In a directed graph, each edge has a direction, meaning that it connects a source node to a target node. In contrast, an undirected graph has no direction, meaning that edges simply connect pairs of nodes without specifying a source or target. Additionally, each edge can be associated with a weight that represents the strength or importance of the relationship it represents. Specifically, Let $v_i \in V$ denotes a node and  $e_{i j}=(v_i, v_j) \in E $ denote an edge pointing from $v_j$ to $v_i$. The neighborhood of a node $v$ is defined as $N(v) = {u \in V | (v, u) \in E}$. The adjacency matrix A is a $n * n$ matrix with $A_{i j} = 1$ if $e_{i j} \in E$ and $A_{i j} = 0$ if $e_{i j} \notin E$. A graph may have node attributes X and edge attributes $X_e$, where $X \in R_{n d}$ is a node feature matrix with $x_v \in R_d$ representing the feature vector of a node $v$ and $Xe \in R_{m,c}$ is an edge feature matrix with $x_{v,u} \in R_c$ representing the feature vector of an edge $(v, u)$.

\subsection{Graph Construction From Spatiotemporal Data}
\noindent  As introduced in the above section, a graph is a data structure that consists of nodes and edges. Nodes represent entities in the space, while edges represent connections or relationships between those entities. Edges can be directed or undirected, indicating the directions of the connection or relationship. An adjacent matrix stores the connection between node pairs. Converting spatiotemporal data into a graph structure involves setting nodes and edges from the spatiotemporal data. Only a small amount of spatiotemporal data has an explicit graph structure, and most spatiotemporal data lack a prior graph structure. We need to determine the latent graph structure from spatiotemporal data through node setting and edge setting as discussed in the following subsections.

\begin{figure}[!h]
     \centering
        \includegraphics[scale=0.10]{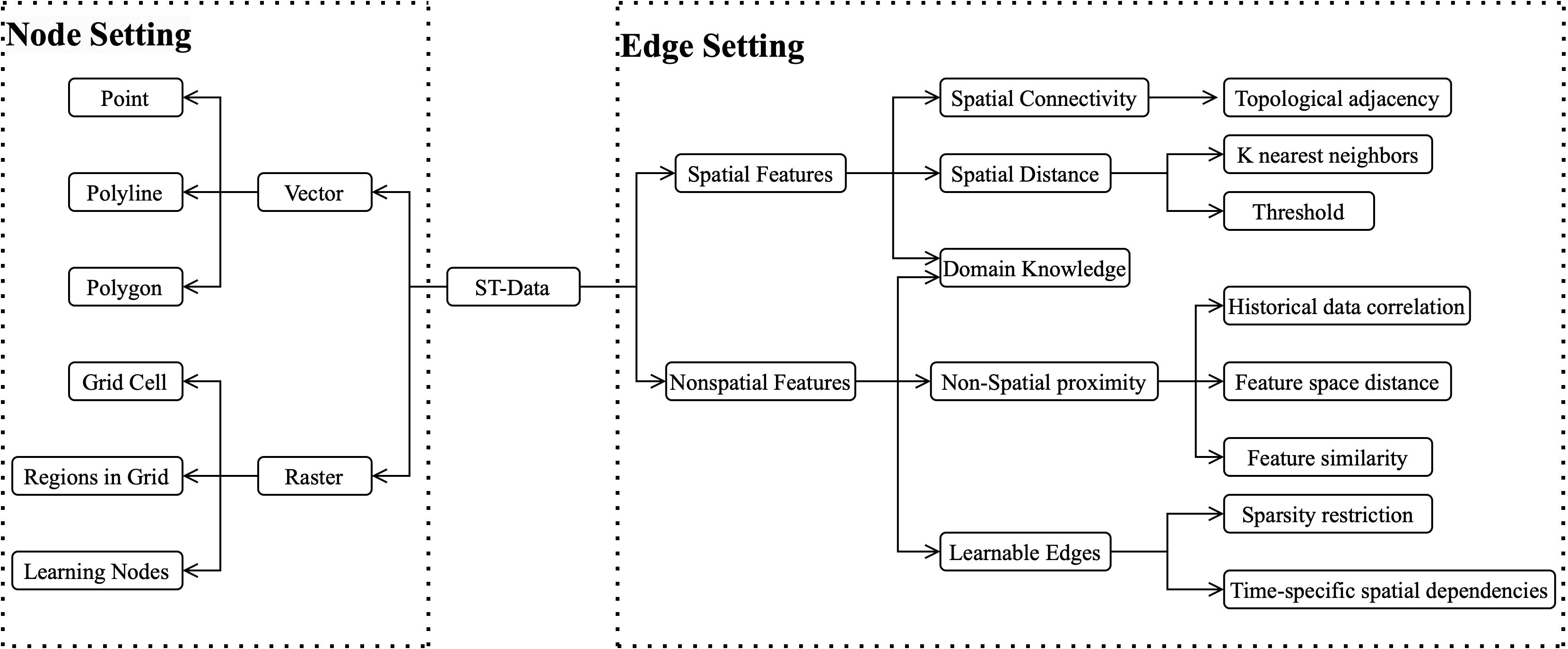}
      \caption{Graph construction from spatiotemporal data.}
        \label{fig:graphcons}
        \vspace{-8mm}
\end{figure}

\subsubsection{Node setting}

Regardless of spatiotemporal data types such as spatiotemporal event, spatiotemporal trajectory, spatiotemporal raster, spatiotemporal data can be categorized into two types from the spatial aspects, including vector data and raster data. 

\textit{\bfseries Vector data}: Converting vector to graph shares similarities among points, polylines, and polygons. For point data such as environmental monitoring sensors \cite{yu2022deep}, EEG sensors \cite{zhang2019learning}, points can be regarded as nodes in a graph, and attributes associated with the points can be assigned as node features. Converting polyline and polygon data to graph is similar to point datasets, the difference is that nodes in the graph can represent line and polygon in the original datasets in addition to points, such as road segments in road network \cite{lu2020agstn, fang2020constgat}, geographic units delineated from terrain polygon approximation \cite{zeng2022graph}, cities extracted from administration units \cite{wang2020pm2}, skeletal joints between bones \cite{li2018spatio} and sensors located at the corresponding roads of the traffic networks \cite{huang2020lsgcn}.

\textit{\bfseries Raster data}: Raster data, which doesn’t possess an explicit graph structure, can be converted to a graph by treating each grid cell as a node in the graph. However, the number of grid cells grows rapidly as the number of rows or columns of raster data increases, especially when the raster data represents a large space with high spatial resolution. An alternative way is to preprocess raster data to regions of interest and then build the graph from these intermediate regions. This can significantly reduce the number of nodes in the graph, making it more computationally efficient and easier to analyze. Xiang et al., created a flow direction map from the precipitation grid using d8 algorithm and then convert the water flow direction map into a directed graph that represents a watershed for rainfall-runoff modeling \cite{xiang2021high}. Zeng et al. converted DEM grid to TIN grid and constructed a GNN constrained by environmental consistency for  landslide susceptibility evaluation, in which the TIN cells aids terrain polygon approximation for environmental consistency evaluation since it can reflect environmental attributes such as slope of the geographic units \cite{zeng2022graph}.  Saueressig et al. partitioned MRI images to supervoxels using Simple Linear Iterative Clustering (SLIC) algorithm and discarding supervoxels that lie outside the brain volume. A graph is then constructed from the remaining supervoxels to identify tumor morous volume for automatic brain tumor segmentation \cite{saueressig2022joint}. Bhattacharya et al. extracted skeletal graph per frame from an RGB video of an individual walking to classify perceived human emotion \cite{bhattacharya2020step}. In addition to using pre-trained object detectors or fixed and predefined regions to extract graph nodes, some researchers learn nodes dynamically from raster data. For example, Duta et al. proposed a model that learns nodes attach to well-delimited salient regions without using any object-level supervision. They found that these localized and adaptive nodes are well correlation with objects in the video \cite{duta2021discovering}.

\subsubsection{Edge setting\label{edge}}
Attributes in the spatiotemporal data that impact information flow between nodes can be set as edge features, such as direction between source node and sink node, wind directions of source node in an air quality monitoring network \cite{wang2020pm2}. The adjacent matrix that stores the connectivity information among nodes usually derives from spatial and non-spatial features in terms of spatial connectivity, spatial distance, non-spatial proximity, learnable proximity, and domain knowledge as shown in Figure \ref{fig:graphcons}.  

\textit{\bfseries Spatial connectivity}:
Spatial connectivity indicates whether two nodes are directly connected or not \cite{park2020st}. It is usually measured by topological adjacency between nodes. For example, \cite{fang2020constgat} defined the road network as a directed graph, in which nodes are road segments, and an edge exists between two nodes if two road segments share the same junction. \cite{li2018spatio} defined the connected edges between joints if body bones connect two joints. Raster data can be converted to a graph by treating each grid cell as a node in the graph and connecting each node to its 8 adjacent cells in the grid \cite{geng2019spatiotemporal}. This results in a graph with a regular structure, where each node has 8 edges. In addition to the spatial neighborhood, geographically distant but conveniently reachable regions can be correlated \cite{geng2019spatiotemporal}.  For example, transportation connectivity plays an important role when performing spatiotemporal prediction in a large spatial scale. This kind of spatial connectivity could be induced by roads like motorway, highway or public transportation. \cite{geng2019spatiotemporal} defined edges between distant regions if they are directly connected by these kinds of roads.

\textit{\bfseries Spatial distance and directions}:
Spatial distance measures distances such as Euclidean distance between two nodes. The k nearest neighbors of a node or distances between two nodes that are smaller than a threshold \cite{meng2021cross} are the two common strategies used to set up edges between nodes. In the former one, edges link a node and its k nearest neighbors. In the latter one, if two nodes are close enough, which is less than the predefined threshold, we put an edge between two nodes. For example, when constructing a graph from in-situ monitoring stations, \cite{mao2021hybrid} established a threshold of 200 km between two air quality monitoring stations to determine whether the two nodes should be connected or not . \cite{multi_attentionN} computed the pairwise road network distance between traffic sensors in a road network to construct the road network graph . For the distance between regions, the distance between the centroids of two regions is computed. As for the more accurate version, the connectivity edge can be defined as whether the distance between two region boundaries is smaller than a certain threshold if region shapes are given \cite{du2019beyond}.  Spatial distance and directions can also be combined with spatial connectivity to measure the spatial proximity between two nodes. \cite{park2020st} computed the proximity information using connectivity, the distance, and direction of the edges between two connected nodes to predict the future traffic speed at each sensor location .

\textit{\bfseries Non-spatial proximity}:
Nonspatial proximity measures the similarity between node pairs with non-spatial features, such as the Euclidean distance of nodes in the feature space, Pearson correlations of historical records. For example, in landslide susceptibility evaluation, \cite{zeng2022graph} defined the edge in the graph structure as the environmental relationships between nodes, in which a consistent environment measured by the similarity of environmental factors such as drainage, stratum, and soil types is a basis for node connections. In some scenarios, it might be inappropriate to measure the distance between two nodes only using the geolocation features (i.e., Latitude and Longitude) since features can be significantly different even if two nodes are fairly close. For example, the distance between stations in the Elysian Park and Downtown LA is less than 2 miles, however, the territorial characteristics are significantly different. Furthermore, the different characteristics (e.g., Tree fraction or Impervious fraction) can affect weather observations (especially, temperature due to urban heat island effect). Thus, considering only physical distance may improperly approximate the meteorological similarity between two nodes \cite{seo2018automatically}. When making predictions for a region, it is intuitive to refer to other regions that are similar to this one in terms of functionality.\cite{geng2019spatiotemporal} defined the edge between two vertices as the POI similarity. \cite{fang2021spatial} utilize both spatial neighbors and semantical neighbors of road nodes to consider spatial correlations comprehensively.

\textit{\bfseries Learnable weight matrix}:
To address the problem of lacking known graph structure, learnable weight matrix has been applied in some research to automatically learn the graph structure during the training process. The weighted adjacent matrix can be set as one of the learning targets instead of manually assigning the connections among nodes, making the graph more interpretable and computationally efficient. For example, \cite{cachay2021world} set each cell in a global Sea Surface Temperature (SST) grid as a node in a graph and enforce a sparse connectivity structure through learning to detect and characterize SST teleconnections for seasonal forecasting up to six months ahead. \cite{li2019classify} applied a learnable mask with the adjacent matrix of a complete graph to learn the dynamic latent graph structure. \cite{feng2020dynamic} introduced a dynamic graph learning block for correcting the global adjacency matrix with sparse spatial dependency constraints. Besides, although spatial positional correlations among nodes can be utilized to construct a graph, it may not be able to capture the functional correlations between nodes in several learning tasks. Take the electrode graph for example, domain scientists are more interested in the latent functional graph in which the edge describes the correlation between the function of different parts of the brain instead of the correlation between spatial positions of electrodes \cite{li2019classify}, thus, \cite{li2019classify} utilized the adjacency matrix of a complete graph and learnable weight matrixes to learn the latent functional graph structure of the brain to overcome the issue of having no prior knowledge of a proper graph and the authors have found remote but functional correlated areas in the brain. In addition, when a GNN model uses pre-defined edge weights based on distance or similarity measures, the static graph structure can be limited in capturing the dynamic interactions among nodes over time. A learnable weight matrix enables learning the evolution of node spatial dependencies. \cite{han2021dynamic} proposed a dynamic graph construction method to learn the time-specific spatial dependencies of road segments for traffic speed forecasting.

\textit{\bfseries Domain knowledge}:
Domain knowledge can also be applied to optimize the construction of a graph in combination with the other four edge-setting strategies, for example, by assisting in choosing a reasonable threshold when building edges based on spatial distances between nodes. Take building a graph from air quality sensors for example, Wang et al. identified a set of critical domain knowledge for PM2.5 forecasting and proposed a knowledge-enhanced GNN by explicitly encoding domain knowledge into the attribute matrices and graph structure. Specifically, the wind speed of the source node, the wind direction of source node, the direction from source node to the sink node, distance between the source and the sink are incorporated into edge features. A geographical knowledge constraint was also applied to build the adjacency matrix, that is, most aerosol pollutants are distributed near the ground and mountains lying along two cities will hinder the PM2.5 transport of the pollutants. Thus, the authors set that the pollutants can transport from one city to another only if the distance between two cities is less than 300 km and the mountains between them are lower than 1200 m \cite{wang2020pm2}. Bao et al. built a physics-guided graph model in which the graph structure is driven by the partial differential equations that describe the underlying physical processes to improve the prediction of water temperature in river networks, enabling capturing the dynamic interactions among multiple segments in a river network \cite{bao2021partial}.

\section{Spatial-temporal Graph Neural Network Techniques}

  
            
\begin{figure}[ht]
\vspace{-2mm}
\begin{centering}
\includegraphics[scale=0.13]{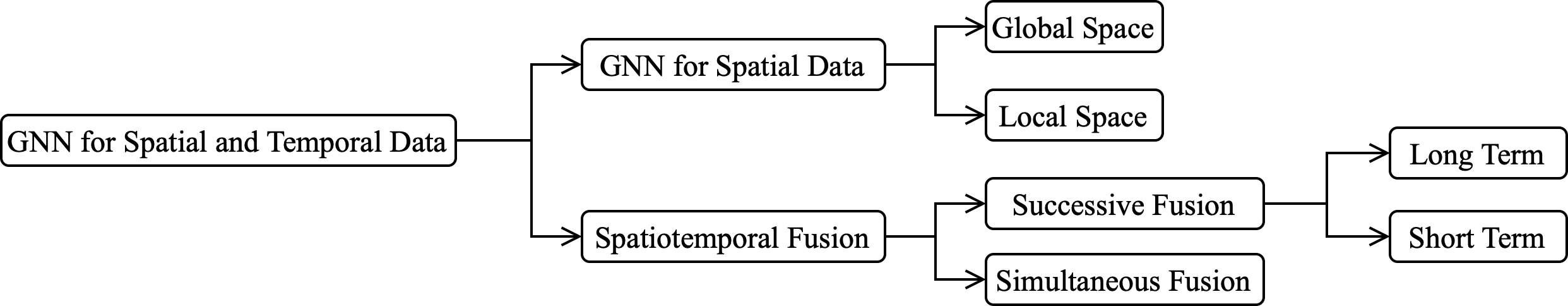}
\vspace{-3mm}
\caption{Taxonomy of spatial-temporal graph neural network techniques }
\vspace{-4mm}
\label{fig:taxonmoy}
\end{centering}
\end{figure}

\noindent   This section focuses on the taxonomy and representative techniques utilized for each category and subcategory of GNN for spatiotemporal data. Based on the relationships learned by the model, the techniques for STGNN can be broadly categorized into GNN for spatial data (Spatial GNN) and GNN for spatiotemporal data (ST-GNN), as shown in Figure \ref{fig:taxonmoy}.
Specifically, GNN for spatial data models static spatial relationships that don't change over time, while ST-GNN captures dynamic relationships over space and time.
The aforementioned two types of GNNs are further categorized based on how they capture spatial and temporal relationships during model training.

\subsection{GNN for spatial data}
\noindent  GNN for spatial data focus on learning the spatial embeddings of nodes or graph for downstream tasks. Typically, these models are composed of functions, represented by Equation \eqref{eq:spatialgnn}, that perform the following operations: (1) calculate the message between the target and source node connected by an edge, (2) integrate messages from all neighbors of the target node, and (3) update node features using the integrated information with a nonlinear transformation. The choice of these functions determines the specific variants of GNNs that can be applied in various applications.

\vspace{-2mm}
\begin{equation}\label{eq:spatialgnn}
\begin{aligned}
    & \boldsymbol{m}_u^{(k)} = \phi_{msg}^{(k)}(\boldsymbol{h}_u^{(k-1)}, \boldsymbol{h}_v^{(k-1)}, \boldsymbol{e}_{u,v}^{(k-1)}),
\end{aligned}
\begin{aligned}  &\boldsymbol{a}_v^{(k)}=\varphi_{agg}^{(k)}\left(\left\{\boldsymbol{m}_u^{(k)}:u \in \mathcal{N}(v)\right\}\right), 
\end{aligned}
\begin{aligned}
    &\boldsymbol{h}_v^{(k)}= \delta_{upd}^{(k)}\left(\boldsymbol{h}_v^{(k-1)}, \boldsymbol{a}_v^{(k)}\right)
\end{aligned}
    \end{equation}
    where $\delta_{upd}, \phi_{msg}, \varphi_{agg}$  represent \textit{node updating}, \textit{message construction} and \textit{message aggregation}.

 GNN methods for spatial data can be categorized into two types based on the way they learn spatial relationships: GNN for local space and GNN for global space. GNN for local space is a neighborhood-based model that focuses on the message passing of each node and its surrounding neighbors, often using methods such as graph convolutional neural networks (GCN) to learn the features between nodes. Compared to GNN for global space, local space GNNs have lower computational complexity and can more quickly process some local information, such as node classification, edge prediction, community discovery, and graph embedding tasks. However, when processing global information, local space GNNs may lose some important information because they ignore the global structure and relationships. GNN for global space is a model based on the entire graph structure, which learns more comprehensive feature representations by considering the interaction between nodes as a whole. Compared to local space GNNs, global space GNNs can model global information and are suitable for large graphs and complex structures.
 To ensure that each branch of the taxonomy is distinct and encompasses non-overlapping techniques, we include spatial modules that can function independently as a GNN model, even if it is originally proposed as a component of a spatial-temporal GNN. This allows readers to select and use modules according to their specific needs.

\subsubsection{GNN for local space}
GNN for local space use a subset of nodes as the neighbors of a target node and calculate a localized representation for each target node. As a result, the aggregating function in Equation \ref{eq:spatialgnn} can be further constrained as follows:

\vspace{-2mm}
\begin{equation}
\boldsymbol{a}_v^{(k)}=\varphi_{agg}^{(k)}\left(\left\{\boldsymbol{m}_u^{(k)}: u \in \mathcal{N}(v), |\mathcal{N}(v)|<|V|\right\}\right) 
\end{equation}
A GNN for local space can be beneficial for processing large graphs and capturing local patterns in the data. However, if the patterns of interest are non-local, using a GNN for local space can limit the model's expressiveness. Therefore, the extent of locality is often chosen based on the task and the properties of the data.It's worth noting that our discussion has focused on the locality of a single layer. However, by stacking multiple layers, GNN for local space can capture global information.

Spatial-based GNN and Approximated Spectral GNN are two GNNs for local space. The former is a neighborhood-based model that learns node feature representation by propagating signals to the node and its neighboring nodes. It is computationally efficient and well-suited for handling local information tasks like node classification and edge prediction. Nevertheless, it may sacrifice vital information when dealing with global information. The latter is a model based on graph spectral theory that uses the graph's Laplacian matrix to propagate signals and learn features. It excels in capturing global information and is ideal for analyzing large and intricate graph structures. However, it comes at the expense of increased computational complexity.

\textit{\bfseries Spatial-based GNN}: 
Spatial-based GNNs define a node's neighborhood using spatial attributes such as distance and contiguity, which is an intuitive way to learn spatial embeddings.
\begin{equation}
    \mathcal{N}(v)= \varphi_{neighbor}(s_v, S),
\end{equation}
GraphSAGE \cite{graphsage} samples neighboring nodes and aggregates messages to learn spatial information inductively. 
Most of the common GNNs with the message-passing scheme fall into this category. 
where $\mathcal{N}_k(u)$ denotes the set of neighbor nodes of $u$ and 
$\mathbf{h}^k_{\mathcal{N}(u)}$ is the embedding of node $u$ after aggregation operation. GAT \cite{GAT} learns the relative weights of neighbors and assigns larger weights to important nodes using the self-attention mechanism.  DCNN \cite{diffusion} defines graph convolutions as a diffusion process in which a probability transition matrix is calculated from the adjacency matrix. Embeddings from multiple diffusion steps are concatenated together as the output. DCRNN \cite{diffusion2} utilizes the bi-directional random walks on the traffic graph to model spatial information. 

\textit{\bfseries Approximated Spectral GNN}: 
Spectral GNNs assume graphs to be undirected and rely on the eigendecomposition of normalized graph Laplacian matrix to define the graph convolution. These convolutions map the graph signal into an orthonormal space where the basis is formed by eigenvectors of the normalized graph Laplacian. Consequently, they are typically global methods. However, many follow-up works approximate the convolution filters and consider only nodes within a certain distance from the central node, resulting in spatial locality. This distance is determined by the adjacency matrix derived from the spatial locations of nodes in the graph.
\vspace{-2mm}
\begin{equation}
    \begin{aligned}
    &  \mathcal{N}(v)= \varphi_{neighbor}(k,A),
    &A=\varphi_A{(S)}
\end{aligned}
\end{equation}
ChebNet is one of the most widely used Spectral GNN techniques.ChebNet is proposed by Defferrard et al.\cite{2016Convolutional} based on the theory that the convolution filter can be approximated by a truncated expansion of Chebyshev polynomials \cite{hammond2011wavelets}.
\begin{equation}
    \begin{aligned}
    &  \widetilde{L}= {2}/{\lambda}_{max}\mathbf{L}-\mathbf{I}_N,
    & \mathbf{g}_w * x = \Sigma^{K}_{k=0} w_k \mathbf{T}_k ( \widetilde{L})x,
    \end{aligned}
\end{equation}
where $g_w$ denotes the convolution filter, $\widetilde{L}$ is the normalized graph Laplacian, ${\lambda}_{max}$ is the largest
eigenvalue of graph Laplacian, $T_k (x)$ denotes the Chebyshev polynomials up to $k^{th}$ order, $\mathbf{I}_N$ denotes the identity matrix and $w_k$ denotes a vector of Chebyshev coefficients. ChebNet approximates the convolution filter by Chebyshev polynomials of the normalized diagonal matrix of eigenvalues.
By limiting the degree of the Chebyshev polynomials used in the convolution operation, we can restrict the size of the receptive field, i.e., the range of nodes that each filter considers. For example, if we limit the degree to k=1, then the filter will only consider the central node and its immediate neighbors. If we increase the degree to k=2, the filter will consider nodes that are two hops away from the central node, which increases the receptive field.
By controlling the degree of the Chebyshev polynomials, we can ensure that the filters in a ChebNet are spatially local and only consider a limited range of nodes. GCN \cite{GCN} introduces a first-order approximation of ChebNet.

\subsubsection{GNN for global space}
GNN for global space use the entire set of nodes in a graph as the neighbors of a target node, allowing for the calculation of a global representation for each target node. The aggregating function in Equation \ref{eq:spatialgnn} can be further constrained as:
\begin{equation}
\boldsymbol{a}_v^{(k)}=\varphi_{agg}^{(k)}\left(\left\{\boldsymbol{m}_u^{(k)}: u \in V \right\}\right) 
\end{equation} 
As the global message consists of all pairwise messages between nodes' embeddings, the computation cost is a $O(n^2)$. GNNs for global space have an advantage in some cases where the spatial relationships between the nodes are not relevant or when dealing with non-Euclidean graphs that do not have a natural spatial embedding. However, it can also limit the expressiveness of the model if the patterns of interest are local or require explicit spatial reasoning. 

\textit{\bfseries Complete Graph}: Many works consider every pairwise distance between two nodes and build a complete graph to capture the spatial dependencies \cite{yuSTGCN}. 
While most GNN models use predefined adjacency matrices to learn spatial embeddings by aggregating messages between nodes, many work to learn hidden spatial correlations between nodes and automatically learn adjacency matrices from data. We can differentiate STGNN methods based on whether the pairwise proximity between nodes is learned, i.e. adaptive, or predefined. For example, \cite{photopower} first calculates the distance between two nodes using the haversine formula, followed by assigning edges using  a Gaussian kernel and a threshold value. Another example is \cite{TAMP-S2GCNETS}, which learns a normalized self-adaptive adjacency matrix by considering the pre-defined “cost” staying in the same node based on the learnable node embedding. They introduce a hyperparameter $d_{uu}$  as the “cost” of staying in the same node $u$ to guarantee the entry of the matrix is larger than 0, and use the matrix as the normalized Laplacian. Towards more effective and robust learning of both spatial and spectral characteristics, they represent graph diffusion as a matrix power series. As this process can be completed in the data preprocessing step and don't require further adjustment, it doesn't fall under the category of spatial-temporal GNN methods. Such adaptive graph learning methods may fix some potential graph construction's inadequacy, which is caused by the failure to take into account implicit spatial correlations and the lack of domain knowledge.

Some literature suggest that some STGNN adaptive graph learning method rely on randomly initialized learnable matrices. MTGNN and Graph Wavenet are two classical examples of two typical adaptive graph learning methods based on randomly initialized learnable matrices. MTGNN \cite{wulearnableA} calculates the adjacency matrix from learned node embeddings
\begin{equation}
    \begin{aligned}
        & E_s = \tanh(\omega N_1 \Theta_1),
    \end{aligned}
    \begin{aligned}
        & E_t = \tanh(\omega N_2 \Theta_2),
    \end{aligned}
    \begin{aligned}
        & A=ReLU(tanh(\omega(E_s E'_t-E_tE'_s))),
    \end{aligned}
\end{equation}
where $N_1, N_2$ are randomly initialized node embedding, $\Theta_1,\Theta_2$ are parameters, $\omega$ is a hyper-parameter that controls the saturation rate of activation function. Graph WaveNet \cite{Graphwavenet}  uses an adaptive matrix that leverages node embedding to capture hidden spatial dependencies to improve performance. The adaptive matrix is learned end-to-end and formulated as $A=Softmax(ReLU(E_s E_t^T))$. While this approach is quite like data preprocessing, it is categorized as a spatial-temporal GNN model because the node features are learned from data.

\textit{\bfseries Spectral GNN}: 
In the early stages of GNN, the Fourier transform is used to translate the spatial domain graph signal into the spectral domain, where the convolution computation may be performed\cite{2020Graph}.Based on this, the graph convolution operation is defined as:
\begin{equation}
\begin{aligned}
       g * x = F^{-1}(F(g)\odot F(x)) = U(\mathbf{U}^T g \odot \mathbf{U}^T x )
\end{aligned}
\end{equation}
where $\odot$ represents convolution operator and $\mathbf{U}^T g$ denotes filter in the spectral domain. The majority of spectral GNNs concentrate on improving the computation process. Using eigenvectors and eigenvalues of the graph Laplacian matrix, spectral GNNs operate in the spectral domain.
\vspace{-1mm}
\begin{equation}
\begin{aligned}
    & x *_G g_{\theta} =\mathbf{U} g_{\theta} \mathbf{U}^T x,
\end{aligned}
\begin{aligned}
    & g_{\theta}=diag(\mathbf{U}^T g),
\end{aligned}
\begin{aligned}
    &\widetilde{L}=\mathbf{U} \mathbf{\Lambda}\mathbf{U}^T,
\end{aligned}
\begin{aligned}
    &\widetilde{L}=\mathbf{I}_N-\mathbf{D}^{-1/2}\mathbf{A}\mathbf{D}^{-1/2},\\
\end{aligned}
\end{equation}
where $\widetilde{L}$ is the normalized graph Laplacian matrix, $\mathbf{D}$ is the diagonal degree matrix, and $\mathbf{I}_N$ is the identity matrix, $\mathbf{\Lambda}$ is the diagonal matrix of eigenvalues (spectrum). Traditional spectral GNNs are global methods because they operate on the entire graph at once, without directly considering the spatial locality of the nodes. While some GNN variants such as graph convolutional networks (GCNs) or ChebNet use localized filters to capture spatial locality, original spectral GNNs use global spectral information to represent the graph structure. Specifically, spectral GNNs convert the node features into the spectral domain, where they are multiplied by a learnable weight matrix. This transforms the features into a new space where the graph structure is represented by the eigenvectors and eigenvalues of the Laplacian matrix. After the multiplication, the features are transformed back to the original space using the inverse Laplacian transform.

\textit{\bfseries Spatial Attention}: 
Bringing attention to the GNNs is being actively studied. One use of attention is to compute the score between the target and source node and scale the importance of source neighbor nodes when doing message aggregation \cite{GAT}. When the attention is calculated on any node pairs in the graph, the model is a GNN for global space, 
\begin{equation}
\begin{aligned}
    \boldsymbol{a}_v^{(k)}  =\varphi_{agg}^{(k)}\left(\left\{\alpha_{u,v}\boldsymbol{m}_u^{(k)}: u \in V \right\}\right)
\end{aligned}
\end{equation}
GAT\cite{GAT} incorporates the attention mechanism into node
aggregation to consider the
importance of neighbor nodes in spatial dependencies learning. The operation is defined as:
\begin{equation}
\begin{aligned}
\mathbf{a}^{t+1}_v & = \rho(\sum_{u\in \mathcal{N}_u}\alpha_{uv}Wa_u^t),
\end{aligned}
\begin{aligned}
\alpha_{u,v}  & =\frac{\exp(LeakyReLU(\boldsymbol{at}^T[\boldsymbol{Wa}_v\parallel\boldsymbol{Wa}_u]))}{\sum_{k\in \mathcal{N}_u} \exp(LeakyReLU(\boldsymbol{at}^T[\boldsymbol{Wa}_v\parallel\boldsymbol{Wa}_k]))},
\end{aligned}
\end{equation}
where $\alpha_{u,v}$ is the attention score between nodes $u$ and $v$, $W$ is the weight matrix associated with the linear transformation for each node, and $\boldsymbol{at}$ is the weight parameter for attention output.  The Latent Correlation Layer in \cite{caolearnAbyattention} uses the self-attention mechanism to learn the adjacency weights. The input $X \in \mathbb{R}^{N\times T}$ is fed into a Gated Recurrent Unit (GRU) layer to calculate the hidden state corresponding to each timestamp  $t$ sequentially. The last hidden state is used to represent the entire time series and calculate the adjacency weight matrix $W$ by the self-attention mechanism, $Attention=Softmax(QK'/\sqrt{d})$, where $Q$ and $K$ denote the representation for Query and Key and calculated by linear projections with learnable parameters in the attention mechanism; and $d$ is the hidden dimension size of $Q$ and $K$. The overall time complexity of self-attention is $O(N^2d)$.

Attention may be applied to both the adjacency matrix and the graph Laplacian. The attention-bassed STGCN (ASTGCN) \cite{ASTGCN} learns a spatial attention matrix and does the Hadamard product with the Chebyshev polynomials to get the adjusted polynomials.
GMAN \cite{multi_attentionN}  defines the spatial attention between two nodes based on the dot-product of the concatenation of the spatial-temporal embedding and the hidden embedding of the two nodes.
Simeunovi{\'c} et al. \cite{PVpower_lstm_transformer} incorporate a slight number of modifications to the transformer architecture. First, they apply three 1D-convolutional layers to the input sequence along the time axis to extract three variations of the input features. Then, they use Chebyshev GCN to create the Query, Key, and Value matrix. Finally, the attention mechanism is applied to the three matrices to get the spatial embedding of each node. 
\subsection{Spatial-temporal GNN}
\noindent Spatial-temporal GNNs focus on capturing both the spatial and temporal dependencies to address spatial-temporal correlated tasks. Spatial-temporal GNNs are neural models that capture the dependence of graphs via message passing between the nodes of graphs. A plethora of sophisticated spatial-temporal GNN structures with massage passing mechanisms have been proposed to capture temporal correlations and spatial dependencies \cite{2020Graph}. 
Based on how spatial-temporal relationships are learned, spatial-temporal GNN methods can be categorized into two types: 
1) Successive Spatial-temporal GNN: The spatial and temporal patterns are learned via individual modules and integrated together as the spatiotemporal relationship; and 2) Simultaneous Spatial-temporal GNN: A GNN-based model that takes the spatial relationship and temporal information as input and jointly learns spatial-temporal embeddings.In other words, Successive Spatial-temporal GNN propagates information sequentially through time, whereas Simultaneous Spatial-temporal GNN processes spatial and temporal information simultaneously in a single layer. Successive Spatial-temporal GNN has the ability to construct time series models to manage graph data and performs well in tasks that factor in time, but each time step requires a large computation. Simultaneous Spatial-temporal GNN has the advantage of efficiently processing spatial and temporal information in a single layer, but it is not appropriate for all tasks, such as those requiring time series analysis.




\subsubsection{Successive Spatial-temporal GNN} 
Individual modules are designed for modeling spatial and temporal relationships, respectively. Existing methods typically rely on graph convolutions for node embedding learning and temporal learning modules, modeling structural and attribute similarity. 
At each layer, the spatial and temporal representations can either be concatenated or kept separate until the final output layer, which is usually a fully connected network. \cite{geng2019spatiotemporal} learns the spatial and temporal dependencies sequentially. Studies like DCRNN \cite{diffusion2},  STGCN \cite{yuSTGCN}, and ASTGCN \cite{ASTGCN} use two separate components to capture temporal and spatial dependencies, respectively. DCRNN \cite{diffusion2} utilized graph convolutional networks for spatial-temporal prediction. Specifically, it employs a diffusion graph convolution network to describe the information diffusion process in spatial networks. In addition, it uses RNN to model temporal correlations, similar to ConvLSTM. STGCN \cite{yuSTGCN} uses CNN to model temporal correlations.  ASTGCN \cite{ASTGCN} uses two attention layers to capture the dynamics of spatial dependencies and temporal correlations, It performs convolutions among recent, daily, and weekly components. These methods directly capture spatial dependencies and temporal correlations. They then feed the spatial representations into temporal modeling modules to indirectly capture the spatiotemporal influence.
According to how the temporal information is modeled, this category can  be further grouped into two subcategories: 1) Successive Short-term Spatial-temporal GNN; and 2) Successive Long-term Spatial-temporal GNN.
\vspace{-1mm}
\begin{equation}
    G^{t+1} = \varphi(\varphi_{spatial}, \varphi_{temporal},G^t) ,
\end{equation}

\textbf{\textit{Successive Short-term Spatial-temporal GNN}}:

Short-term spatial-temporal GNNs capture short-term dynamics by learning the changes in graph representations across neighboring time steps. 
According to how the temporal information is modeled, Short-term spatial-temporal GNNs can  be further grouped into two subcategories: 1) RNN-based methods; and 2)1D-CNN based methods.RNN-based STGNN uses RNNs to handle time series data and is better at handling variable-length sequences and dependencies within sequences. However, it suffers from potential gradient problems when dealing with long sequences. On the other hand, 1D-CNN based STGNN uses one-dimensional CNNs (1D-CNNs) to capture patterns in time series data, with better computational efficiency and ability to handle long sequences. However, it cannot handle variable-length sequences or dependencies within sequences as well as RNN-based STGNN. 

\begin{itemize}[fullwidth]
  \item \textit{\bfseries RNN-based Methods}:
  RNNs process the time information in an  autoregressive manner by retaining past information in hidden states. However, since the next state in RNN models depends only on the current state, it is biased towards the most recent inputs in the sequence, resulting in the older inputs having less influence on the current output. In addition, RNN's key shortcoming, such as gradient disappearance and gradient explosion, are difficult to address during training\cite{2009Reservoir}. To address these limitations and increase the time range that models can learn, gated models such as LSTM and GRU models are introduced. In recent works, RNN models and graph convolutions are often combined, with one category of such work replacing the matrix multiplication in the RNN model with a graph convolutional operation \cite{PVpower_lstm_transformer}. In the classical LSTM cell, the cell state $c$ and the output $h$ are updated recursively from the input sequence $x$ using gating operations involving matrix multiplications. In graph-convolutional long short term memory network (GCLSTM) cells \cite{PVpower_lstm_transformer}, the cell state and output are updated by replacing the matrix multiplications with spectral graph convolutions,
\begin{small}
\vspace{-3mm}
\begin{equation}
\begin{aligned}
f_t= & \sigma\left(W_f *_{\mathcal{G}} X_t+U_f *_{\mathcal{G}} h_{t-1}+w_f \otimes c_{t-1}+b_f\right), 
& i_t= & \sigma\left(W_i *_{\mathcal{G}} X_t+U_i *_{\mathcal{G}} h_{t-1}+w_i \otimes c_{t-1}+b_i\right) \\
c_t= & f_t \otimes c_{t-1} +i_t \otimes \tanh \left(W_c *_{\mathcal{G}} X_t+U_c *_{\mathcal{G}} h_{t-1}+b_c\right) ,
& o_t= & \sigma\left(W_o *_{\mathcal{G}} X_t+U_o *_{\mathcal{G}} H_{t-1}+w_o \otimes c_t+b_o\right) \\
h_t= & o_t \otimes \tanh \left(c_t\right),
\end{aligned}
\vspace{-2mm}
\end{equation}
\end{small}
where $\sigma$ is the sigmoid function, $W_f *_{\mathcal{G}} \cdot$ is the graph convolution and $\otimes$ is the Hadamard product. 

Owing to the presence of numerous gated units, the computational cost of LSTM is rather considerable. Thus, GRU compresses the gated units of LSTM into update gate and reset gate. $u_t$ symbolizes the update gate, which decides how to merge the information of the current input time step with the memory of the previous time step and $r_t$ denotes the reset gate, which determines the amount of memory reserved from the previous time step to the current time step. The computation procedure of GRU is defined as follows:
\begin{equation}
\begin{aligned}
& u_t=  \sigma\left(W_u \centerdot x_t+U_u * C_{t-1}+ b_u\right),  
& r_t=  \sigma\left(W_r \centerdot x_t+U_r * C_{t-1}+ b_r\right)\\
& \widetilde {C_t} =  \tanh \left(W_C \centerdot x_t+U_C \centerdot (C_{t-1}\odot r_t)+ b_C\right), 
& C_t =  u_t \centerdot C_{t-1} + (1-u_t)\centerdot \widetilde {C_t}& 
\end{aligned}
\end{equation}

Another example of GRU-based temporal learning network is T-GCN\cite{zhao2019t}. This model uses a recursive structure to study the spatial and temporal correlation. At each time step, GCN and GRU work in sequence to learn spatial and temporal relationships from graph signals. Each GCN and GRU in this model's computation may be written as:
\begin{equation}
    \begin{aligned}
       & f(X,A)  =   \sigma(AXW_0), 
       & u_t  =  \sigma(W_u[f(A,X_t),h_{t-1}+b_u]), \\
       & r_t   =   \sigma(W_r[f(A,X_t),h_{t-1}+b_r]), 
       & c_t =  \tanh(W_c[f(A,X_t),(r_t*h_{t-1})]+b_c)\\
       & h_t =  u_t * h_{t-1} + (1-u_t) * c_t
    \end{aligned}    
\end{equation}
where $f(A,X_t)$ is the output of spatial GCN at time step t and was utilized by the GRU to obtain the hidden state at the same time.In STGNN, it is possible to embed RNN-based temporal learning networks into GNN-based spatial learning networks. DCRNN\cite{diffusion2} is an example of STGNN with such an architecture since it incorporates a diffusion GCN into the GRU network to capture the spatial and temporal dependencies.
It also uses the encoder-decoder architecture to deal with multi-step prediction. Specifically, DCRNN encodes the input sequence into a fixed-length vector and predicts the output for the next T steps. The matrix multiplications in the GRU are replaced by the diffusion convolution operation defined by the diffusion GNN: 
\begin{equation}
    \begin{aligned}
       & \mathbf{r}^{(t)} = \sigma (\Theta_r * \mathcal{G}[X^t, H^{t-1}]+b_r), 
        & \mathbf{u}^{(t)} = \sigma (\Theta_u * \mathcal{G}[X, H^{t-1}]+b_u), \\
        & \mathbf{C}^{(t)} = \tanh (\Theta_C * \mathcal{G}[X^t, (r^{(t)}\centerdot H^{t-1})]+b_c), 
        & \mathbf{H}^{(t)} =  \mathbf{u}^{(t)}  \centerdot H^{(t-1)} + (1-\mathbf{u}^{(t)}\centerdot C^{(t)}), 
    \end{aligned}    
\end{equation}
where $\Theta_R * \mathcal{G}$ is the graph convolution operator with parameter $\Theta_R$.

\item \textit{\bfseries 1D-CNN-based Methods}:
Many spatio-temporal modeling tasks have used RNN-based temporal learning networks, but the recurrent structure of RNN-based models require the sequences to be calculated at each time step, which significantly increases the calculation burden and reduces model efficiency. This problem may be solved by combining 1D-CNN with temporal networks. 

CNN based methods tackles temporal dependencies by applying different kernels to the same time series, however, for capturing global correlations in long sequences. CNN-based methods require stacking multiple layers, which can lead to loss of local information as the dilation rate increases.
STGCN \cite{yuSTGCN} and GraphWaveNet \cite{Graphwavenet} models employ 1D convolution along the time axis to capture temporal depedencies. STGCN uses 1D causal convolution with gated linear units (GLU) to model temporal dynamic, while each spatial convolution module alternates between two 1D convolution layers. specifically, two TCNs and one GCN are stacked to form a ST-Conv block that learns spatiotemporal dependencies in a parallelized manner, i.e., it receives all information of a given time window length as input simultaneously. In mathematical form, the calculation of each ST-Conv block can be defined as:
\vspace{-1mm}
\begin{equation}
    \begin{aligned}
        v^{l+1} = \Gamma^l_1 * ReLU(\Theta^l *_G (\Gamma^l_0 * \tau_{v^l}))
    \end{aligned}
\end{equation}
where $\Gamma^l_0,\Gamma^l_1$ denotes the upper and lower temporal convolution kernel within an ST-Conv block l, $\Theta^l$denotes the spectral kernel of graph convolution. 

Compared to the RNN-based models, the 1D-CNN module can improve the training efficiency with fewer parameters. STGCN also employs an iterative strategy for traffic prediction, where predictions from previous iterations are used for the next iteration, accumulating prediction errors over time.
Similarly, the temporal convolution layers in \cite{photopower} also use 1D convolutions with a one-dimensional kernel filter followed by a sigmoid-gated linear unit to introduce non-linearity. The sigmoid function chooses relevant elements from the input for capturing complex structures and variances in the time series. Another example of combination of 1D-CNN with temporal convolution networks is the Gated-TCN\cite{dauphin2017language}, whose caculation process is defined as:
\vspace{-1mm}
    \begin{equation}
        \begin{aligned}
            F(x) = \tanh (\theta_1 * x)\centerdot  \sigma\left(\theta_2 * x\right)
        \end{aligned}
    \end{equation}
where both $\theta_1$ and $\theta_2$ are learnable parameters in 1D-CNNs, $*$ signifies convolution operation,$\centerdot$ denotes element-wise multiplication mechanism, and $\sigma\left(\theta_2 * x\right)$ represents the gating unit that control the utilization rate of historical information.

Although 1D-CNN is an efficient parallel neural architecture, causal correlations in temporal learning are not modeled in some cases. Typically, in traditonal neural networks, the connections between neurons in each layer are fully connected, which violates the fundamental constraint of time series, as the output of the front (previous time steps) neurons are fully connected to the input neurons of the back (future time steps). The mask mechanism is applied to partially eliminate the layer-by-layer links in the networks while retaining the links from previous time steps to future time steps. To learn short-range to long-range temporal dependencies, 1D-CNN is combined with dilated factors\cite{yu2015multi}. The 1D-CNN with dilated factors can be calculated:
\vspace{-1mm}
\begin{equation}
        \begin{aligned}
            F(s) = (x*_df)(s)\Sigma_{n=i}^{k-1} f(i).x_(s-d\centerdot i)
        \end{aligned}
\end{equation}

Another example is Graph WaveNet (GraphWN) \cite{Graphwavenet}, which combines GCN and stacked dilated causal convolutions at multiple granular levels to model the temporal correlations. A limitation of CNN-based methods is that the large number of kernels required to capture dependencies among all possible combinations of time steps in a time series. This becomes impractical because a kernel of size k would only learn relationships between k time steps. As the length of the input series increases, the number of possible combinations of time steps  grows exponentially, making it challenging to capture long-term dependencies.
\end{itemize}
\vspace{-2mm}
\textbf{\textit{Successive Long-term Spatial-temporal GNN}}: Although LSTM/GRU and stacked 1d-CNN based STGNN models can retain information over time, this information has the tendency to fade and dilute over long sequences. Successive Long-term Spatial-temporal GNN models overcome this issue by learning temporal information from a global perspective, addressing the locality problem.
\vspace{-2mm}
\begin{itemize}[fullwidth]
    \item \textit{\bfseries Transformer-based Methods}: Transformers were originally introduced in the field of machine translation to allow for parallel computation and reduce the negative impact of long dependencies on performance. Unlike traditional RNNs, transformers process time series as a whole, rather than step by step, thereby avoiding recursion.
    Specifically, a GNN extracts node or graph embeddings at each timestamp. These embeddings are then fed into a transformer using a self-attention module to compute similarity scores between any two elements of the time sequence \cite{brain_trans}. 

    Transformers do not rely on past hidden states to capture dependencies with the previous timestamps. Instead, each step has direct access to all other steps, thereby eliminating the risk of losing past information. Moreover, multi-head attention and positional embeddings  provide information about the relationship between different times, enabling transformers to capture complex temporal dependencies in a highly efficient manner.

    Although transformers \cite{PhysicsMeshTemporalAttention} offer powerful modeling capabilities, their autoregressive nature can make the sequential prediction process time-consuming during inference. The sequential procedure of transformers \cite{PhysicsMeshTemporalAttention} may still be time-consuming during inference due to their autoregressive structures. Han et al. \cite{PhysicsMeshTemporalAttention} propose applying a transformer model to latent graph embeddings, with attention over the entire sequence, enabling the model to predict the latent graph embedding for the next step. They learn a simulator that can predict the sequence $(z_1,...,z_T )$ autoregressive based on an initial state $z_0$ and a special system parameter token $z_{-1}$. Then the transformer model, $trans()$ predicts the subsequent latent graph vectors in an autoregressive manner: $\tilde{z}_1 = trans(z_{-1}, z_0), \tilde{z}_t = trans(z_{-1}, z_0, \tilde{z}_1,..., \tilde{z}_{t-1})$. Specifically, at step $t$, the prediction $\tilde{z}_{t-1}$ from the previous step issues a query to compute attention $a_{(t-1)}$ over latents $(z_{-1}, z_0, \tilde{z}_1,..., \tilde{z}_{t-1})$.
    \vspace{-2mm}
    \begin{equation}
        a_{(t-1)}=\text{softmax} \left( {{z}_{t-1}^T W_1^T W_2}/{\sqrt{d'}} [z_{-1}, z_0, \tilde{z}_1,..., \tilde{z}_{t-1}]\right)
    \end{equation}
    Here $W_1$ and $W_2$ are learnable parameters, and $d'$ denotes the length of the vector $z$. Next, the attention head computes the prediction vector
    \begin{equation}
        g_{(t)}=W_3 [z_{-1}, z_0, \tilde{z}_1,..., \tilde{z}_{t-1}]a_{(t-1)}^T
    \end{equation}
    with the learned parameter $W_3$. The multi-head attention model uses $K$ parallel attention heads with separate parameters to compute $K$ vectors $(g_t^{(1)},...,g_t^{(K)} )$ as described above. These vectors are concatenated and fed into an MLP to predict the residual of $\tilde{z}_{t}$ over $\tilde{z}_{t-1}$.
    \begin{equation}
        \tilde{z}_{t}=\tilde{z}_{t-1} + \text{MLP}(\text{concat}(g_t^{(1)},...,g_t^{(K)} )).
    \end{equation}
    This model can propagate gradients through the entire sequence and use information from all previous steps to produce predictions. In addition, transformers can simultaneously capture short-term and long-term temporal information when combined with RNN-based temporal networks. For instance, DMVST-VGNN\cite{jin2022deep} combines TCN and Transformer for long-short range temporal learning, whereas Traffic STGNN\cite{wang2020traffic} achieves multi-scale temporal learning through multi-network integration by employing GRU for learning short-term temporal dependencies and Transformer for learning long-term temporal dependencies.
\vspace{-1mm}
\end{itemize}
\subsubsection{Simultaneous Spatial-temporal GNN}


This category of works \cite{fang2021spatial,ASTGCN,traffic2} learn the dependencies between spatial and temporal dimensions simultaneously. Capturing complex localized spatial-temporal correlations simultaneously can lead to effective prediction, as it allows us to model the spatial-temporal data at a fundamental level.
\begin{equation}
    G^{t+1} = \varphi_{spatiotemporal}(G^t) ,
\end{equation}
\begin{itemize}[fullwidth]
    \item \textit{\bfseries Spatial attention-based Methods} Spatial attention-based models calculate the similarity of input entries and mix them by the similarity after softmax operation. Standard transformer-like models calculate the similarity based on transformed features, the key idea of using the attention mechanism in \cite{2.5+1D} is to use a similarity defined by spatiotemporal proximity of graph nodes as characterized by its (2.5+1)D scene graph. For two nodes $v_1, v_2 \in V$, a similarity kernel $\kappa$ is: 
    \begin{equation}\label{eq2.5+1}
        \kappa(v_1,v_2|\sigma_S,\sigma_t)= \exp \left( - {\|p_{v_1}-p_{v_2}\|^2}/{\sigma_S^2} - {\|t_{v_1}-t_{v_2}\|_1}/{\sigma_T^2}\right),
    \end{equation}
    which captures the spatial-temporal proximity between $v_1, v_2$ for scales $\sigma_S$ and $\sigma_T$ for spatial and temporal cues, respectively. Then, the (2.5+1)D-Transformer is given by:
    \begin{equation}
        F=\text{softmax} \left(K(V,V|\sigma_S,\sigma_T)\right) V_F,
    \end{equation}
    where it uses $K$ to denote the spatiotemporal kernel matrix constructed on $V$ using Equation \ref{eq2.5+1} between every pair of nodes, $V_F$ is the Value matrix transformed from the node embeddings. Such a similarity kernel merges feature from nodes in the graph that are spatiotemporally nearby.
\item \textit{\bfseries Temporal Attention-based Methods}:
The attention mechanism is employed to assign weights to the edges in a graph according to their temporal distance. These weights are determined by evaluating the similarity between the node features connected by each edge at various time steps. This enables the model to focus on neighboring nodes that possess the most pertinent temporal information for the current prediction task. GMAN \cite{multi_attentionN} defined the temporal attention between two time steps at a specific node as the dot product of the concatenation of the spatial-temporal embedding and the hidden embedding of the node at the two time steps.

\item \textit{\bfseries Spatiotemporal Graph}:
Spatio-temporal data can be represented as a matrix $X \in \mathbb{R}^{N\times T}$, where $N$ denotes the number of locations and $T$ is the length of time stamps. Each row corresponds to a time series for a spatial node, while each column represents a spatial signal at a particular time stamp. Traditionally, a single adjacency matrix $A \in \mathbb{R}^{N\times N}$ is created for each timestamp to capture the spatial relationships among different junctions of the network at a particular time. However, it is important to consider the temporal information in defining connections. Specifically, there are three types of influences in a spatiotemporal network. First, nodes can directly influence their neighbors at the same time step, reflecting spatial dependencies. Second, nodes can directly influence themselves at the next time step due to temporal correlations in the time series. Lastly, nodes can even directly influence their neighbor nodes at the next time step due to synchronous spatial-temporal correlations. These three types of influences arise because information propagation in a spatiotemporal network occurs simultaneously along the spatial and temporal dimensions.  Note that the index of spatial information is arbitrary as each spatial location can be associated with a vertex on the spatial graph, and the edges will provide information about the relative position of the nodes. We can reshape the matrix to form a vector $x$ of length $NT$, where the element $x_{(s;t)}:= x_{(s-1)T+t}$ is the feature value corresponding to the $s-th$ vertex at time $t$. Various connection schemes have been proposed to construct a spatiotemporal graph.

To capture the complex spatiotemporal relationships from different timestamps, USTGCN \cite{USTGCN} introduces cross-spacetime edges which enable the model to capture spatiotemporal dependencies between nodes across different timestamps. Specifically, at timestamp $t$, each node aggregates the traffic features of ego (target node) and its neighbors from the previous 1 to $(t - 1)$ timestamps. 
The product graph \cite{STGST} built a spatiotemporal graph by constructing a spatial graph $G_s = (V_s; E_s; A_s)$ with $|V_s| = N$, reflecting the graph structure in the spatial domain and a temporal graph $G_t = (V_t; E_t; A_t)$ with $|V_t| = T$, reflecting the graph structure in the temporal domain. A product graph, denoted as $G = Gs \Diamond Gt = (V; E; A)$ can be constructed to unify the relations in both spatial and temporal domains,  allowing us to process data jointly across space and time. The product graph has $NT$ nodes and an appropriately defined adjacency matrix $A$. The operation $\Diamond$ interweaves two graphs to form a unifying graph structure. The edge weight $A_{(s1;t1);(s2;t2)}:= A_{(s1-1)T+t1;(s2-1)T+t2}$ characterizes the relation, such as similarity or dependency, between the $s1-th$ spatial node at the $t1-th$ time stamp and the $s2-th$ spatial node at the $t2-th$ time stamp. Three commonly used product graphs are Kronecker product, Cartesian product, and strong product which is a combination of Kronecker and Cartesian products. There is also a small part of work that  generates adjacency matrix by using other mechanism. DGCRN\cite{li2023dynamic} synchronously generates the dynamic adjacency matrix of each time step through a recurrent adaptive graph learning mechanism. This recursive mechanism is based on the hidden states to capture spatial information and build the graph structures for each time step. DSTAGNN\cite{lan2022dstagnn} proposed a self-attention-based adaptive graph learning method to establish the connection between the graph structures and the hidden states.


\end{itemize}

\section{Applications}
\subsection{Transportation}
\noindent   With the advent of intelligent public transportation systems, we can now benefit from predictive traffic flow and travel time. However, most existing models ignore global and local spatiotemporal dependencies, leading to inaccurate predictions and suboptimal routing. GNN-based models can capture spatiotemporal dependencies and dynamics of transportation systems to improve traffic management and enhance overall transportation efficiency.
Most models dedicated to predicting traffic are validated in the PeMS dataset and the METR-LA dataset as shown in Table \ref{tab:table1}. 

\subsubsection{Transportation Systems' Supply and Demand Prediction} The ability to predict transportation system supply and demand is crucial for traffic management, optimization, and planning. \cite{zeng2023combining} proposed a split-attention relational GCN that constructed the metro system as a knowledge graph from historical Origin-Destination (OD) matrix to explore spatiotemporal correlations. \cite{makhdomi2023gnn} used an attention mechanism in a GNN-based framework for OD passenger flow prediction, in which both linear and non-linear dependencies in passenger requests from different places are preserved. \cite{jin2023spatio} proposed a multi-time multi-GNN with a Gated CNN framework and a multi-graph module to capture spatial dependencies. \cite{zhang2022comprehensive} built a graph convolutional and comprehensive temporal neural network that combined the Transformer and LSTM to capture global and local temporal dependencies and used a GCN framework to capture spatial features of the subway network. \cite{lu2021dual} proposed a dual attentive GNN to predict passenger flow distribution, using both inbound and outbound graphs and a multi-layer graph spatial attention network to capture dependencies between inbound and outbound flows. GNN-based models are commonly used to handle non-euclidean spatial data but typically rely on predefined graphs. To overcome this limitation, \cite{yu2021deep} proposed a Neural Relational Inference method that generates an optimal graph by leveraging external data with a Variational Auto Encoder (VAE)-based model. The study also used bayesian priors to encode domain knowledge on generated graphs and explored the interpretability of GNNs' connections for spatio-temporal predictions in the transport domain.

\subsubsection{Traffic Anomalies Detection} 
Intelligent traffic networks are increasingly popular in smart cities. An important role of GNN in traffic systems is to model transportation systems and road networks to detect anomalies. An attention-based temporal GCN was proposed in \cite{wu2022lane} to detect lane changes at the road segment level. To improve traffic anomaly detection in IoT-based intelligent transportation systems (ITS), \cite{wang2022contrastive} proposed a GNN-based model that utilizes contrastive learning to improve the learning of latent representations for downstream tasks. They also introduced a graph augmentation approach for learning different views of the data.

\subsubsection{Idle vehicles relocation/repositioning} 
On-demand ride services like taxis and transportation network companies (TNCs) are popular in transportation. Service providers face the challenge of relocating idle vehicles to improve order rejection rates, passenger waiting time, vehicle occupancy rates, and driver revenues. \cite{yu2021deep} developed a GNN-based learning framework to represent road networks as graphs and improve repositioning decision-making by maintaining road network structure and learning dependencies in data.

\subsection{Human mobility}
\noindent  Human mobility data is crucial in various domains, including smart transportation and epidemic modeling. There are several sub-fields within the domain of human mobility, with trajectory prediction and people-flow prediction being the most popular ones that employ GNN. 
				
\subsubsection{Trajectory Prediction} 
GNN models are popular for location prediction and trajectory recovery due to their ability to capture periodicity, regularity, and transitional information in human trajectories. \cite{sun2021periodicmove} proposed a GNN-based neural attention model for recovering lengthy and sparse human trajectories. They constructed a directed graph for each trajectory to preserve transitional information and used attention mechanisms to capture multi-level periodicity and shifting periodicity in human trajectories. \cite{dang2022predicting} proposed a Graph Convolutional Dual-attentive Networks (GCDAN) for trajectory reconstruction. They used a bidirectional diffusion graph convolution to capture spatial dependencies and employed a dual-attentive mechanism via a Sequence to Sequence mechanism to capture transitional information and location correlations.\cite{terroso2022nation} proposed a single model to process all the mobility areas via GNN. The inter-urban displacements with larger spatial granularity are anticipated by considering the latent relationships among large geographical regions. \cite{wang2020using} developed a multi-pattern passenger flow prediction framework (MPGCN) by constructing a sharing-stop network to capture passengers’ relationships. A clustering framework is proposed to preserve the mobility patterns hidden in bus passengers, and based on the mobility patterns, a GCN2Flow framework is proposed for passenger flow prediction.

\subsubsection{People-flow Prediction} People-flow prediction estimates the number of people entering or exiting a zone to reflect real-time travel demand. It plays a significant role in urban planning, emergency response, and urban roadway operations\cite{chen2019data}. (1)Tourist flow prediction.
GNN models have been used to address challenges in existing models which focus only on local or regional levels and static spatial connections. \cite{saenz2023nation} modeled nationwide tourist mobility as a graph and used heterogeneous tourism data to conduct tourist inflow forecasting as an edge prediction task. \cite{zhou2023graph} proposed a graph-attention-based framework that embeds dynamic spatial connections into a weight-dynamic multi-dimensional graph with a node attribute sequence. Their graph-attention network can model explicit and implicit dynamic spatial connections and learn high-dimensional spatiotemporal features for tourist flow prediction. (2) Crowd flow prediction. It estimates the inflow and outflow of crowds in a city. However, traditional grid-based prediction ignores irregular regions and distant correlations. To address these problems,\cite{li2022crowd} proposed a GNN-based model for crowd flow prediction in irregular regions. The model uses CNN and GNN to capture micro and macro spatial dependencies and includes a location-aware and time-aware graph attention framework to capture inter-region correlations. \cite{xing2022stgs} developed a GNN-based method that jointly constructs spatial and temporal graphs from grid maps to capture correlations among distant regions.

\subsection{Point Of Interest (POI) Recommendation}
\noindent   POI recommendation is a personalized location-based service that suggests places of interest to users based on their preferences or interests. GNN models have gained significant attention in this area, as they can effectively represent relationships between POIs as graphs. There are two main types of POI recommendations: social and geographical. 

\subsubsection{Social POI recommendations}In Social POI recommendation uses users' activities on social media to recommend potential friends or products. \cite{salamat2021heterographrec} introduced HeteroGraphRec, a GNN for spatial data model that represents the social network as a heterogeneous graph and leverages attention mechanisms to effectively aggregate information from multiple sources. The model considers users and items as nodes and establishes connections between them to obtain detailed information about the user's preferences. Social POI recommendation models also incorporate the time dimension, such as user sessions and chronological orders of their activities. \cite{kefalas2018recommendations} developed a successive spatial-temporal fusion model using a hybrid tripartite graph that combines seven separate unipartite and bipartite graphs with spatial elements like the user, location, and temporal elements like sessions. Similarly, \cite{dai2021personalized} proposed a model that learns representations for users and POIs by jointly modeling their relations, sequential patterns, geographical influence, and social ties in a heterogeneous graph. The model then personalizes sequential patterns for users to provide personalized POI recommendations.\cite{wang2022spatiotemporal} developed a simultaneous-fusion Spatio-temporal GNN (SGNN) for session-based recommendation. The model aims to replicate users' activity patterns from a spatiotemporal perspective to predict their next click. SGNN comprises two modules: Spatiotemporal Session Graph (SSG) and Preference-aware Attention (PAN). The SSG module models all session sequences to simulate possible user activity patterns, while the PAN module divides possible interests into users' session-wide interests and current interests. This approach enables the model to effectively recommend the next click based on the user's spatiotemporal behavior.\cite{min2021stgsn} proposed STGSN, a successive spatial-temporal GNN model for modeling social networks. The model utilizes an embedding approach to capture the spatial properties of nodes for each time slice, and an attention-based strategy to aggregate the graphs from a temporal perspective. 

\subsubsection{Geographical POI recommendation} Geographical POI recommendation leverages users' trajectories to predict locations they may be interested in. The SpatioTemporal Aggregation Method, STAM, incorporates temporal ordering of one-hop neighbors into neighbor embedding learning to provide spatiotemporal neighbor embeddings\cite{yang2022stam}. It allows for the creation of spatial-to-spatiotemporal aggregation approaches through simultaneous spatial-temporal fusion GNN. Luo et al. introduced Spatio-Temporal Attention Network(STAN), a  simultaneous spatial-temporal GNN for suggesting locations based on a user's trajectory \cite{luo2021stan}. STAN uses self-attention layers to explicitly incorporate spatiotemporal information from all check-ins along a route, enabling direct interaction between nonadjacent sites and nonconsecutive check-ins. \cite{yuan2020spatio} built a SpatioTemporal Dual Graph Attention Network(STDGAT) that concurrently describes the dynamic context and sequential user behaviors. A dual graph attention network was introduced to describe global query-POI interaction and time-varying user preferences on destination POIs. \cite{li2021discovering} proposed the Sequence-to-Graph (Seq2Graph) augmentation method for POI recommendation. Seq2Graph augmentation is an iterative process that propagates collaborative signals by correlating POIs from different sequences. They also introduced the Sequence-to-Graph POI Recommender (SGRec), which simultaneously learns POI embeddings and infers a user's temporal preferences from the graph-augmented POI sequence. SGRec maximizes collaborative signals for user preference modeling.

\subsection{Climate and Weather}
\noindent  Weather refers to atmospheric conditions at a specific place and time, while climate describes long-term weather patterns in a specific area. Both weather and climate forecasting are important to prepare for upcoming weather conditions and understand long-term changes in Earth's climate.Several scientists have attempted to develop spatial-temporal models to investigate and forecast weather and climate conditions with the support of GNN. 

\subsubsection{Weather} \cite{peng2021cngat} introduced a categorical node graph attention network (CNGAT), a GNN model that fuses spatial and temporal information simultaneously for improved Radar quantitative precipitation estimate (RQPE). The precipitation estimation area was partitioned into subareas that were treated as nodes to form an input graph for CNGAT. All nodes were then categorized according to the temporal mean radar reflectivity for precipitation estimation with an attention mechanism. \cite{wilson2018low} proposed WGC-LSTM, which uses graph convolutions to capture spatial relationships and integrates them with LSTM to consider spatial and temporal relationships simultaneously. In response to wind speed forecasting challenges caused by the stochastic and highly varying nature of wind,\cite{khodayar2018spatio} proposed a Graph Convolutional Deep Learning Architecture (GCDLA). This model represents wind farms as nodes and uses edges to depict neighborhood relationships based on mutual information in historical data. The graph structure leverages strong spatial correlations in wind speed and direction data, while an LSTM network extracts temporal features. GCDLA accurately predicts wind speed using these spatial-temporal features.\cite{keisler2022forecasting} built a simultaneous spatial temporal fusion model to predict atmospheric conditions for several days ahead. The model uses a bipartite network to convert latitude and longitude into icosahedron grid nodes, which are connected by edges to form a graph. GNN is employed to learn latent features for each node. The resulting icosahedron grid latent features are then transformed back into atmospheric features in latitude/longitude grid cells for forecasting.\cite{lira2021frost} introduced GSTA-RGC, a model that simultaneously fuses spatial and temporal information to forecast the lowest temperature in a local experimental field for a specified number of hours. The model employs a 3D tensor that contains nodes, temperature, humidity, and time, as well as a 2D matrix that includes nodes' geographical information to construct GNNs. GAT is used to extract spatial-temporal similarity features from the 3D tensor and 2D matrix for the final prediction.

\subsubsection{Climate} \cite{cachay2021world}
proposed Graphiño, a GNN for global space for seasonal forecastings, such as predicting El Niño-Southern Oscillation (ENSO). The model creates an initial graph with grid cells as nodes, and edges are learned based on the connectivity between locations. The graph representation is then fed into a fully connected multi-layer perceptron for seasonal climate forecasting. \cite{ni2022ge} proposed the Graph Evolution-based Spatio-Temporal Dense Graph Neural (GE-STDGN) network to address challenges in long-term spatio-temporal forecasting, including high feature redundancy, long-term prediction dependency, and complex spatial relations between geographical locations. The model conducts correlation analysis to preserve useful features while removing those with internal and prediction result in relevance, building a spatial-temporal graph with these features. Graph Evolution (GE) is introduced to improve the model's node association analysis by deriving an optimal adjacency matrix based on the selected features. 

\subsection{Disaster Management}
\noindent   Climate change increases the prevalence and severity of natural hazards such as hurricanes, wildfires, floods, and droughts. Extreme natural events interacting with exposed and vulnerable humans can lead to disasters. Early detection and forecasting of natural hazards can help us better prepare to mitigate hazard effects by improving early warning systems with rich spatial and temporal information. An extensive growing number of studies have applied GNN for natural disaster management including preparedness,prevention, and response.

\subsubsection{Natural disaster preparedness}
Anticipating unseen disasters improved the public readiness for extreme disasters. (1) Floods: \cite{sit2021short} used StreamGConvGRU, a simultaneous spatial temporal fusion model, to predict streamflow in the next 36 hours at locations with monitoring sensors. The model employed three Graph Convolutional Gated Recurrent Unit Networks for discharge and precipitation measurements from the previous 36 hours and anticipated precipitation in the next 36 hours. \cite{farahmand2021spatial} introduced an attention-based spatial-temporal graph convolution network (ASTGCN) for urban flood nowcasting at the census tract level, enhancing situational awareness. The model consists of multiple spatial-temporal blocks, each with a spatial and temporal attention module followed by a convolutional graph layer, enabling the capture of dynamic spatial and temporal dependencies in a step-by-step manner.
(2) Landslides: \cite{kuang2022landslide} developed a landslide forecasting attentive GNN (LandGNN), a successive spatial-temporal GNN model, for predicting land displacement by aggregating spatial correlations across monitoring locations using graph convolutions. LandGNN incorporates a locally historical transformer to learn the interdependence of neighboring areas, capturing the growth of local interactions for land displacement prediction. \cite{zeng2022graph} proposed a spatial-based GNN with environmental consistency constraints for landslide susceptibility evaluation. They constructed a graph using geographic units in a triangulated irregular network as nodes and refined node representations by aggregating information from nearby landslides. This approach generates reliable landslide susceptibility evaluation results in complex and heterogeneous environments. (3) Tropical cyclones: \cite{xiang2021high} proposed an Attention-based Multi ConvGRU model (AM-ConvGRU) for typhoon trajectory prediction. This successive spatial-temporal fusion GNN integrates 2D and 3D environmental information to enhance typhoon trajectory prediction. The 2D typhoon features were extracted using a Generalized Linear Model, while 3D features were extracted using Residual Channel Attention Block (RCAB) and Multi-ConvGRU. RCAB selects high-response isobaric planes for future typhoon movement and assists the neural network in analyzing a typhoon's 3D structure.

\subsubsection{Disaster prevention and response}
It aids in mitigating the impact of extreme events on the public.
\cite{yuan2022spatio} proposed a successive spatial-temporal fusion GNN model for predicting road traffic conditions and inundation status during flood events. The model constructs a graph based on road segments using historical traffic speeds and connectivity between segments. The model makes predictions by considering factors such as road segment proximity, elevation differences, and adjacency between segments. \cite{jin2020ufsp} proposed a successive spatial-temporal fusion GNN model, UFSP-Net, which combines CNN, GCN, and RNN models to predict urban fire situations for the next time step to improve situation awareness. A region graph was created using fire records and region boundary data. CNN and GNN extracted spatial representation from Fire Situation Awareness Images (FSAI), while gated recurrent unit networks (GRUN) captured temporal information from image sequences. \cite{chen2021significant} introduced Wavelet GNN for predicting Significant Wave Height (SWH). Meanwhile, \cite{ghosh2022unsupervised} proposed a Global and Local GNN (GLEN) to classify social media posts for disaster management with limited labeled data. GLEN combines unweighted local and weighted global graphs, with the local graph preserving per-instance context in token representations and the global graph learning domain-agnostic feature representation. The representations from both graphs are combined and fed into a bi-directional LSTM network for text classification.

\subsection{Environment}
\noindent   Environmental science is crucial for understanding natural systems and their interaction with human activities. Machine learning has emerged as a powerful tool to support environmental science, as it allows us to analyze large amounts of data to model and predict environmental phenomena, such as air pollution concentration. GNN has been widely applied in studying wireless sensor networks for environmental monitoring. These advanced technologies help environmental scientists create more effective environmental management and protection strategies.

\subsubsection{Environmental assessment and prediction} Environmental assessment and prediction enhance our understanding of the surrounding environment. \cite{chen2021group} proposed a Hierarchical Group-Aware GNN (GAGNN) for countrywide city air quality forecasting. This GNN for spatial data learns local and global spatial information by modeling spatial and latent relationships between cities using a city graph and a city group graph. The group correlation encoding module captures dependencies between city groups for nationwide air quality forecasting, regardless of distance. \cite{mao2021hybrid} proposed a Graph Convolutional Temporal Sliding Short-Term Memory model (GT-LSTM) to predict air pollutant concentrations, such as PM2.5 and CO, in urban areas. Combining GCN and LSTM, the model extracts spatial-temporal features simultaneously. GT-LSTM models horizontal pollution transmission using a multi-layer bidirectional LSTM with a temporal sliding strategy. \cite{qi2019hybrid} proposed a hybrid model using Graph Convolutional networks and Long Short-Term Memory networks (GC-LSTM) to forecast hourly-scaled air pollutant concentrations. The authors created an undirected graph with stations as nodes and edges reflecting connections, capturing geographical dependence between stations. LSTM is applied to extract temporal dependency between data at different time spans. \cite{gao2021graph} proposed a Graph-based Long Short-Term Memory (GL-STM) model to predict PM2.5 concentrations for all air quality monitoring stations by introducing a parameter adjacency matrix to learn spatial dependencies between stations. This simultaneous spatial-temporal fusion model eliminates the need to train a separate model for each station to determine a region's PM2.5 concentration. \cite{wang2020pm2} developed PM2.5-GNN, a successive spatial temporal fusion model, to predict PM2.5 concentrations in multiple city clusters for the next 72 hours, considering long-term dependencies between PM2.5 concentrations. The spatial transport of PM2.5 between cities and temporal diffusion is modeled using a graph structure and  RNN, respectively.

\subsubsection{Missing observation data imputation} Environmental sensors are critical in preventing adverse effects on the environment, but malfunctions can cause data to go unrecorded. In response, \cite{chen2022adaptive}  proposed the Adaptive Graph Convolutional Imputation Network (AGCIN) to accurately and efficiently impute missing values in environmental sensor networks by concurrently representing spatial and temporal relationships\cite{chen2022adaptive}. The AGCIN framework combines adaptive graph convolution with a bidirectional gated recurrent unit (GRU) network. Specifically, the framework constructs a spatiotemporal graph by computing similarities between sensor sequences and node embeddings and then uses the resulting adjacency matrix as input for the graph convolution layers. These layers are further integrated with bidirectional GRU to capture spatial and temporal dependencies. By leveraging the temporal dependencies, AGCIN is capable of reconstructing missing values.

\subsection{Power System}
\noindent   Urban development has increased energy usage, especially electricity,in recent centuries.A power grid is a network that delivers electricity from power plants to consumers through transmission and distribution lines, substations, and equipments. It ensures reliable power supply in areas of high demand, such as cities. Applications of GNNs in power grid field includes: (1) predicting power generation, consumption, and transmission, (2) detecting and repairing faults or attacks.

\subsubsection{Spatiotemporal forecasting in power systems}
Precise prediction is crucial for behind-the-meter (BTM) load and photovoltaic (PV) generation. \cite{khodayar2020spatiotemporal} introduced a spatiotemporal BTM load and PV forecasting problem (ST-BTMLPVF), disaggregating residential units' historical net loads and anticipating their unobservable time series. They proposed a simultaneous spatial temporal fusion GNN model to identify net load's key spatiotemporal characteristics using prior BTM load and PV data. 
Researchers have also employed GNNs for PV performance and load prediction. \cite{photopower} developed a spatiotemporal GNN that represents plants as nodes connected based on their geographical distribution and leverages coherence across power plants to predict PV power. \cite{lin2021spatial} proposed a successive spatial-temporal fusion GNN model for short-term electric load forecasting, capturing the spatial dependency between houses in the same or neighboring areas with the self-adaptive Graph WaveNet and temporal dependency with dilated causal convolution. \cite{sun2022fast} designed a temporal and topological embedding deep neural network (TTEDNN) model for predicting transient stability in power grids. The model efficiently forecasts transient stability by extracting temporal features from time-series data and incorporating spatial characteristics from the power grid's structure and electrical parameters.

\subsubsection{Faults detection and repair} Modern power grids are complex systems that are vulnerable to attacks and system failure, making it important to develop methods to detect and repair these issues using GNN. \cite{boyaci2022cyberattack} developed a Chebyshev GCN model for detecting cyberattacks in large-scale AC power systems, taking active and reactive bus power injections as inputs. \cite{liao2020fault} proposed a model improving transformer failure diagnosis precision using GCN's adjacency matrix and graph convolutional layers for complex fault analysis.\cite{liu2021searching} presented an efficient search method for identifying critical cascade failures by modeling the power network using GNN and studying cascading failure mechanisms within a smaller model parameter space. Using gated GNN, \cite{de2021fault} proposed a model for automatic fault localization on distributed networks. The model collects issue data in a graph, with connections and node attributes representing feeder architecture and containing information such as operational devices, electrical characteristics, and point measurements. In addition, high-quality time series data is crucial for essential power grid functions, such as state estimation and distributed energy resource scheduling. However, interruptions in communication can cause metering infrastructure data to be incomplete. To address this, Kuppannagari et al. proposed a Spatio-temporal GNN based Denoising Autoencoder (STGNN-DAE) for imputing missing data in the power grid\cite{kuppannagari2021spatio}. The model leveraged temporal and spatial correlations from the power grid topology and time series data to perform Missing Data Imputation (MDI).

\subsection{Neuroscience }
\noindent  The brain shapes our personalities and our behaviors. Understanding how the brain works can give us a deeper insights into ourselves and help us to improve treatment and diagnose mental illnesses. GNN can model the human brain owing to its ability to capture the spatial dependencies of brain data and represent non-Euclidean data structures commonly observed in neuroimaging data. Applications of GNN in neuroscience can be mainly classified into three categories: mental disease diagnosis, functional brain network construction, and other brain activity prediction/classification.

\subsubsection{Brain disease diagnosis, classification, and treatment}
This direction involves predicting the presence of a particular disease in a given neuroimaging data and is typically performed at the graph level. Medical images play an extremely important role in the process and segmentation, classification, and recognition of medical images are the main tasks in medical imaging. When utilized GNN for specific tasks, image data can be represented as a graph structure appropriate for the use of GNNs. The construction of graphs varies depending on the type of input data. For EEG data, EEG electrodes/channels are often used as nodes in the graph. MRI (Magnetic Resonance Imaging) images are usually partitioned into regions that serve as nodes in a graph and the average time series is often computed for graph construction.fMRI (functional MRI) data usually undergoes preprocessing, such as motion correction,distortion correction, spline resampling, intensity normalization, pre-whitening, and spatial smoothing \cite{qiu2022unrevealing, yan2022modeling}. 

\subsubsection{Brain functional connectivity}
Brain functional connectivity supports understanding the relationship between the structure and function of the brain. It models connections between brain regions and learns representations of the brain \cite{brain_trans}, which provides insights into the underlying mechanisms of various brain disorders. In the realm of learning brain functional connectivity, ABIDE and HCP datasets in Table \ref{tab:table1} are commonly utilized to assess the efficacy of models. Currently, the majority of deep learning models utilize regular images as their input data, GNN has the capability of processing irregular brain connection images. \cite{ktena2018metric} pioneer proposed utilizing GCN for brain connection learning. \cite{yang2019interpretable} constructed a dynamic functional connectivity matrix to predict the state of consciousness of the brain.The result demonstrated the effectiveness of graph convolution methods in predicting cortical signals. \cite{brain_trans} learned dynamic graph representation of the brain connectome with spatiotemporal attention. The model ingested a temporal sequence of brain graphs to obtain the dynamic graph representation with a spatial attention module and a temporal attention model with a transformer encoder successively, which enables incorporating the dynamic characteristics of the functional connectivity network which fluctuates over time.

\subsubsection{Other brain activity prediction/classification}
Tasks such as brain age prediction \cite{stankeviciute2020population}, and sleep stage classification \cite{li2022attention} also play an important role in brain study.
\cite{li2022attention} proposed a hybrid spatiotemporal graph convolutional network (ST-GCN) that combines both dynamic and static components, along with inter-temporal attention blocks. This approach allows for capturing long-range dependencies between electrodes for sleep stage classification. For multichannel EEG emotion recognition, \cite{li2021attention} proposed an ASTG-LSTM (attention-based spatiotemporal graphic long short-term memory) model with spatio-temporal attention mechanisms to maintain invariance against emotional intensity fluctuations. Graph construction can also operate on a different scale than modeling the entire brain image as a graph. For example, population graphs use individual subjects as nodes with both neuroimaging and non-imaging data as node features\cite{stankeviciute2020population}, which operate on a population scale, and neuroimaging data is condensed as node features.

\section{Performance Evaluation}
\noindent  This section provides a comparative analysis of various GNN models across multiple spatiotemporal applications. We gathered experimental data from three traffic flow datasets and two biological datasets. In Table \ref{tab:table1}, a list of publicly accessible benchmark datasets is presented, along with detailed information such as data links and domains.

\begin{table*}[]
\centering
\resizebox{\columnwidth}{!}{

\begin{tabular}{l|ll}
\hline
Dataset  & Link                                            & Application \\ \hline
ABID I     & http://fcon\_1000.projects.nitrc.org/indi/abide/abide\_I.html                             & Biology and Medical \\
HCP s1200  & https://www.humanconnectome.org/study/hcp-young-adult/document/1200-subjects-data-release & Biology and Medical \\\hline
Gowalla  & https://snap.stanford.edu/data/loc-gowalla.html & POI         \\
Foursquare & https://sites.google.com/site/yangdingqi/home/foursquare-dataset                          & POI                 \\\hline
PeMS03   & https://pems.dot.ca.gov/                        & Transportation     \\
PeMS04   & https://pems.dot.ca.gov/                        & Transportation     \\
PeMS07   & https://pems.dot.ca.gov/                        & Transportation     \\
PeMS08   & https://pems.dot.ca.gov/                        & Transportation     \\
METR-LA  & https://github.com/liyaguang/DCRNN              & Transportation     \\
PEMS-BAY & https://pems.dot.ca.gov/                        & Transportation \\ 
NYC taxi & https://www1.nyc.gov/site/tlc/about/tlc-trip-record-data.page & Transportation \\
NYC bike & https://www.citibikenyc.com/sytem-data & Transportation \\
\hline
\end{tabular}
}
\caption{Public benchmark dataset for spatiotemporal GNN models}
\vspace{-8mm}

\label{tab:table1}
\vspace{-2mm}
\end{table*}

\subsection{Traffic Flow Forecasting}
\noindent   GNN models have attracted considerable interest because of their capacity to analyze graph data and capture spatial dependencies between data points. These models have been especially effective in traffic flow forecasting, where the road network inherently possesses a graph structure. In this survey, we examined numerous studies that concentrated on using GNN models for traffic flow forecasting and presented our findings in Table \ref{tab:table3} and Table \ref{tab:table4}.

\subsubsection{Comparison Models}
The performance of STGNN models was evaluated based on their ability to forecast traffic flow. The comparison included models: 

\begin{itemize}[leftmargin=*]
    \item DCRNN: A successive spatial-temporal GNN that captures spatial dependencies using bidirectional random walks and temporal dependencies using RNN with scheduled sampling. 
    \item STGCN: A successive spatial-temporal GNN that sequentially captures temporal information using a 1D-CNN and extracts neighboring spatial features through graph aggregation. 
    \item ASTGCN: A successive spatial-temporal GNN that employs spatial and temporal attention to capture the dynamic spatiotemporal correlations. 
    \item Graph WaveNet: A successive spatial-temporal GNN that combines graph convolution with dilated casual convolution to capture spatial-temporal information.
    \item FC-GAGA: A successive spatial-temporal GNN that captures spatial and temporal information with distinct components. 
   \item STFGNN: A simultaneous  spatial-temporal GNN that learns spatial and temporal information by integrating Spatial-Temporal Fusion Graph  Neural Modules and Gated CNN module. 
  \item StemGNN: A simultaneous spatial-temporal GNN model that applies Graph Fourier Transform and Discrete Fourier Transform jointly in a spectral domain.
  \item STSGCN: A simultaneous spatial-temporal GNN that employs multiple localized spatial-temporal subgraph modules to simultaneously capture the spatial-temporal correlations. 
  \item DSTAGNN: A simultaneous spatial-temporal GNN that utilizes multiple spatial-temporal modules to learn spatial-temporal information.
  \item STGODE: A simultaneous spatial-temporal GNN that utilizes a distinct adjacency matrix to learn spatial-temporal information. 
\end{itemize}
\subsubsection{Benchmark Datasets}
\begin{itemize}[fullwidth]
    \item  PEMS-BAY: A traffic dataset provided by the California Transportation Agency, covering from 01/01/2017 to 05/31/2017, with data from 325 sensors across the Bay Area \cite{li2017diffusion}.
    \item METR-LA: A widely used traffic data collected from highway loop detectors in Los Angeles County\cite{li2017diffusion}, includes data from 207 sensors and covers from 03/01/2012 to 06/30/2012.
    \item PeMS04: A dataset covers the traffic data of San Francisco Bay from 1 January 2018 to 28 February 2018 gathered by 3848 sensors, including speed and occupancy statistics\cite{chen2001freeway}.

\end{itemize}

\begin{table*}[]
\centering
\resizebox{\columnwidth}{!}{
\begin{tabular}{l|llllllllll}
\hline
Dataset  &               & 15 min &        &      & 30min &        &      & 60min &         &      \\ \hline
         & Models        & MAE    & MAPE   & RMSE & MAE   & MAPE   & RMSE & MAE   & MAPE    & RMSE \\ \cline{2-11} 
         & with pre-defined Graph-Structure \\ \cline{2-11} 
METR-LA  & DCRNN & 2.77 & 7.3\% & 5.38 & 3.15 & 8.8\% & 6.45 & 3.60 & 10.5\% & 7.59 \\
         & STGCN         & 2.88   & 7.62\% & 5.74 & 3.47  & 9.57\% & 7.24 & 4.59  & 12.70\% & 9.4  \\
         & Graph WaveNet & \textbf{2.69}   & \textbf{6.90\%} & \textbf{5.15} & \textbf{3.07}  & {8.37\%} & {6.22} & {3.53}  & {10.01\%} & {7.37} \\
         & GMAN & 2.77 & 7.25\% & 5.48  & \textbf{3.07}  & \textbf{8.35\%} & 6.34 & \textbf{3.40} & \textbf{9.72\%} & \textbf{7.21}
         \\ 
         & TransGAT & 2.71 & 7.69\% & 5.20 & 3.15 & 8.70\% & \textbf{6.16} & 3.48 & 9.80\% & \textbf{7.21} \\
         \cline{2-11}
         & without pre-defined Graph-Structure  \\ \cline{2-11} 
         & Graph WaveNet(only adaptive matrix) & 2.80   & 7.45\% &{5.45} & {3.18}  & {9.00\%} & {6.42} & {3.57}  & {10.47\%} & \textbf{7.29} \\
         & FC-GAGA       & {2.75}   & {7.25\%} & {5.34} & {3.1}   & {8.57\%} & {6.3}  & {3.51}  & {10.14\%} & 7.31 \\
         & FC-GAGA(4 layers)       & \textbf{2.70}   & \textbf{7.01\%} & \textbf{5.24} & \textbf{3.04}   & \textbf{8.31\%} & \textbf{6.19}  & \textbf{3.45}  & \textbf{9.88\%} & \textbf{7.19} \\       
         \hline
PEMS-BAY  & with pre-defined Graph-Structure \\ \cline{2-11}
        & DCRNN   & 1.38 & 2.9\% & 2.95 & 1.74 & 3.9\% & 3.97 & 2.07 &  4.9\% & 4.74 \\
         & STGCN         & 1.36   & 2.90\% & 2.96 & 1.81  & 4.17\% & 4.27 & 2.49  & 5.79\%  & 5.69 \\
         & Graph WaveNet & \textbf{1.3}    & \textbf{2.73\%} &  \textbf{2.74} & 1.63  & {3.67\%} & {3.7}  & {1.95}  & {4.63\%}  & {4.52} \\ 
        & GMAN & 1.34 & 2.81\% & 2.82 & \textbf{1.62} & \textbf{3.63\%} & 3.72 & \textbf{1.86} & \textbf{4.31\%} & \textbf{4.32} \\
        & TransGAT &	1.31 & 2.74\%  &	2.78 & \textbf{1.62} &	3.66\% & \textbf{3.69}	& 1.89	& 4.52\%	& 4.38 \\
                 \cline{2-11}
          & without pre-defined Graph-Structure  \\ \cline{2-11} 
         & Graph WaveNet(only adaptive matrix) & \textbf{1.34}   & {2.79\%} & {2.83} & 1.69  & \textbf{3.79\%} & \textbf{3.80} & 2.00 & 4.73\% & 4.54\\
         & FC-GAGA       & 1.36   & 2.87\% & 2.86 & {1.68}  & 3.80\% & {3.80}  & {1.97} & {4.67\%}  & {4.52} \\
          & FC-GAGA(4 layers)       & \textbf{1.34}   & {2.82\%} & \textbf{2.82} & \textbf{1.66}   & \textbf{3.71\%} & \textbf{3.75}  & \textbf{1.93}  & \textbf{4.48\%} & \textbf{4.40} \\
\hline
\end{tabular}
}
\caption{Performance comparison of  spatiotemporal GNN models on PENS-BAY and METR-LA dataset}
\vspace{-8mm}

\label{tab:table3}
\end{table*}

\begin{table*}[]
\centering
\resizebox{\columnwidth}{!}{
\begin{tabular}{l|llllllllll}
\hline
Dataset & Models   & Graph WaveNet  & DCRNN  & STGCN & ASTGCN  & STFGNN & STGODE  & StemGNN  & DSTAGN & STSGCN    \\ \hline
PEMS-04 & MAE & 25.45  & 24.70  & 22.70 & 22.93   & 19.83 & 20.84 & 20.24 &\textbf{19.3}  & 21.19  \\
        & RMSE & 39.70  & 38.12  & 35.55 & 35.22   &  31.88 & 32.82 & 32.15 & \textbf{31.46}  & 33.65  \\
        & MAPE & 17.29  & 17.12  & 14.59 & 16.56   & 13.02 & 13.77  & \textbf{10.03} & 33.65  & 13.90 \\
\hline
\end{tabular}
}
\caption{Performance comparison of spatiotemporal GNN models on PEMS-04 dataset}
\vspace{-10mm}

\label{tab:table4}
\end{table*}

\subsubsection{Evaluation Metrics}
These models are evaluated using metrics mean absolute error (MAE), Root-mean-square error(RMSE) and mean absolute percentage error (MAPE) where $x_i$ is the observed value and $y_i$ is the predicted value.
\begin{small}
\begin{equation}
  \begin{aligned}
        MAE = {(\Sigma_{i=1}^n |y_i-x_i|)}/{n},
    \end{aligned} \quad
    \begin{aligned}
          RMSE = \sqrt{\Sigma_{i=1}^{n}{\Big(({y_i -x_i})/{\sigma_i}\Big)^2} /{n}},
    \end{aligned} \quad
    \begin{aligned}
        MAPE = {\Sigma_{i=1}^{n} ( |y_i-x_i|}/{x_i} ) /{n}
    \end{aligned}
\end{equation}
\end{small}

\subsubsection{Experiment Results}

Table \ref{tab:table3} compares the performance of several spatiotemporal GNN models in predicting 15, 30, and 60 minute traffic flow on the METR-LA and PEMS-BAY datasets. Graph Wavenet outperforms other models in short-term (15min) traffic predictions on both datasets. GMAN, on the other hand, achieves the best performance for long-term (60min) traffic flow prediction and outperforms most models for medium-term prediction. GMAN is recommended for medium- to long-term traffic flow forecasting, while Graph Wavenet is suitable for short-term prediction. GMAN uses a transformer-based attention approach to capture long-term dependencies, while Graph Wavenet employs separate GCN layers to identify local spatiotemporal dependencies effectively. FC-GAGA is a noteworthy model in the table, as it does not rely on prior knowledge of the underlying graph. Its stackable layer architecture allows for the discovery of different graph structures. FC-GAGA performs well in medium- to long-term traffic prediction, suggesting that automatically learning graph structures is crucial. This conclusion is also highlighted by the experimental results presented in Table-4, which compares the performance of various spatiotemporal GNN models using the PeMS04 dataset for traffic prediction. DSTAGNN and StemGNN outperform most models and their superior performance can be attributed to their ability to learn graph structures automatically without prior knowledge.

\subsection{Brain Analysis}
\noindent   Graph models have also been increasingly employed to learn spatial dependencies from raster datasets. This approach has shown great potential in various applications, including the studying the dynamic graph representation of the brain connectome. 
Specifically, the spatiotemporal GNN 
allows researchers to identify functionally related regions in the brain that are located far apart. 
In this survey, we compared the performance of several proposed methods listed in Table 5 on two benchmark brain datasets for analyzing functional brain connectomes.  

\subsubsection{Comparison Models}
The performance of the models listed below was compared. 
\begin{itemize}[fullwidth]
    \item GIN: A Spatial-GNN model.
   \item STAGIN: A Successive Spatial-temporal GNN that combines Transformer and GIN.
    \item GraphSAGE: A Spatial-GNN model that generates low-dimensional representations for nodes. 
    \item STGCN:A successive Spatial-temporal GNN that captures the temporal information with a 1D-CNN and captures the neighboring spatial features with graph aggregation. 
\end{itemize}


\subsubsection{Benchmark Datasets}
The HCP S1200 release is a fMRI dataset that has been widely used in brain studies. Two derived datasets, HCP-Rest and HCP-Task, were created for classification\cite{brain_trans}. 
\begin{itemize}[fullwidth]
    \item HCP-Rest: It contains 1093 preprocessed and ICA-denoised resting-state fMRI data from both male (499) and female (594) participants, with gender labels for classification tasks.
    \item HCP-Task: It consists of fMRI data collected from participants performing activities such as working memory, social and gambling for use in multiple classification tasks.
\end{itemize}

\subsubsection{Evaluation Metrics}

These models are evaluated using accuracy, which represents the proportion of correctly classified samples to the total number of samples and \begin{equation}
    \begin{aligned}
        Accuracy = ({TP+TN})/({TP+FP+TN+FN})
    \end{aligned}
\end{equation}
where $TP$ is a correctly predicted positive example, $TN$ refers a correctly predicted negative example, $FP$ indicates a wrongly predicted positive example, and $FN$ is a wrongly predicted negative example.

\subsubsection{Experiment Results}

As shown in Table \ref{tab:my-table} indicates that STAGIN outperforms GNN for spatial data models and STGCN for gender classification using the HCP-Rest dataset, achieving approximately $5\%$ better accuracy than GIN. STGCN's worse classification accuracy compared to GIN suggests that it does not fully leverage its dynamic capability to capture temporal features in this study. Additionally, STAGIN exhibits superior performance compared to other models for the signal decoding task on the HCP-Task dataset. STAGIN's success on both challenges can be attributed to its capacity to capture spatiotemporal brain characteristics. It achieves this by encoding timestamps with node attributes, using an attention mechanism, and incorporating a transformer encoder. 

\begin{table*}[]
\resizebox{.7\linewidth}{!}{
\begin{tabular}{l|llll}
\hline
    Accuracy & STAGIN & ST-GCN & GIN & GraphSAGE \\ \hline
HCP Rest & 88.20 ± 1.33   & 76.95 ± 3.00 & 81.34 ± 2.40  & 75.48 ± 1.97 \\ 
HCP-Task & 99.19 ± 0.20 & 98.92 ± 0.27 & 93.87 ± 0.66 & 54.52 ± 0.97 \\  
\hline
\end{tabular}
}
\caption{Performance comparison of spatiotemporal GNNs in brain connectome study}
\vspace{-10mm}

\label{tab:my-table}
\end{table*}

\section{Open Challenges and Outlook}
\noindent   Although there have been significant advancements of GNNs in spatiotemporal domain over the past few years, several unresolved issues that require further investigation and promising open research directions remain in this field.
\subsection{Graph Structure Construction from Spatiotemporal Data }
\subsubsection{Building Predefined Graph Structures with Domain Knowledge }
To effectively apply GNN models to spatiotemporal mining tasks, a well-defined graph structure is crucial for efficient information propagation. However, constructing a graph structure solely based on distance may not capture the complete spatial dependencies, which can negatively impact the prediction accuracy. Although researchers have proposed various strategies to support latent graph construction, including building graphs based on spatial connectivity, non-spatial proximity, and domain knowledge (as discussed in Section \ref{edge}), it is important to note tedgehat real-world graph structures are both flexible and complex. With the ever-growing abundance of spatial and temporal data, there are more emerging problems, challenges, and application scenarios that require more powerful latent graph construction methods, especially knowledge-guided latent graph structures. For example, an open research issue in spatial. As an example, one open research issue in spatial-temporal GNNs for traffic forecasting is how to construct an appropriate adjacency matrix with the support of graph theory to represent road traffic networks with greater semantic information \cite{bui2022spatial}.

\subsubsection{Automated Discovery of Graph Structures in Spatiotemporal Data}
The latent graph structure in data is unknown and capturing the latent graph structure from spatiotemporal data is challenging since 1) Positional correlation is not functional correlation, that is, two positionally distant nodes can be functionally close and finding the functional correlation among nodes could improve the performance of GNNs \cite{li2019classify}; 2) In many spatiotemporal application scenarios, spatial dependencies between nodes change over time, and the pre-defined graph structure cannot capture the change \cite{diao2019dynamic, bui2022spatial}. To capture distant but functional dependent nodes among data over time, automatically learning graph structure from spatiotemporal data has become an increasingly active field of research. It is not easy to automatically capture the latent graph structure and a major challenge is that the learned graph must have a meaningful interpretation \cite{du2019beyond}. Incorporating domain knowledge into the graph structure learning process to facilitate meaningful graph structure learning remains a challenging task. One promising research is to construct a physics-guided graph structure that is driven by the partial differential equations describing the underlying physical processes in a river network \cite{bao2021partial}, in which researchers leverage underlying physical phenomena to construct a more meaningful graph. However, there is still much work to be done, as researchers must optimize latent graph structure learning to capture the complexity and richness of real-world systems.

\subsection{Spatial-dependent and time-aware modeling}
\subsubsection{Simultaneous Fusion of Spatial and Temporal Information}
GNNs have emerged as a powerful tool for performing spatiotemporal mining tasks by leveraging the spatial dependencies extracted from the topological structure of graphs. Incorporating temporal correlations can significantly enhance the accuracy of these models. However, effectively integrating spatial and temporal information in GNNs remains a challenge, which requires a capability that goes beyond spatial dependency and considers temporal dependency to predict how the spatiotemporal information might influence the learning target. Current research is predominantly focused on two approaches to integrating spatial and temporal information: simultaneous fusion and successive fusion. Successive fusion employs GNNs alongside other models, such as RNN and LSTM, to separately learn spatial and temporal representations. These representations are then combined for spatiotemporal mining tasks \cite{li2019classify}. On the other hand, simultaneous fusion utilizes specifically designed layers to jointly exploit spatial and temporal representations \cite{fang2020constgat}. A growing trend in spatiotemporal GNN research involves the simultaneous integration of spatial and temporal information, as this approach can capture the dynamic spatial and temporal interdependencies and their potential impact on the learning target. However, this area of research is still in its infancy and lacks comprehensive investigation. Further investigation is required to determine the most effective methods for implementing spatial dependency with GNNs while dynamically capturing a wide range of temporal patterns \cite{chen2020tssrgcn}. 
\subsubsection{Model Local and Global Spatiotemporal Dependencies}
In the realm of spatiotemporal data mining, there exist intricate and dynamic spatial-temporal dependencies between regions. These dependencies encompass both temporal and spatial aspects, with short-term local neighboring fluctuations and long-term global trends in temporal aspects, and local and global correlations in the spatial aspect \cite{feng2020dynamic}.Most existing approaches have primarily focused on modeling the Euclidean correlations among spatially adjacent regions or non-Euclidean pair-wise correlations among possibly distant regions \cite{geng2019spatiotemporal}. However, some spatiotemporal machine learning tasks, such as traffic speed prediction, require consideration of both local and global dynamic spatial-temporal dependencies, as modeling local spatial-temporal dependencies alone ignores the global spatial-temporal corrections \cite{feng2020dynamic}. To explore local and global spatial and temporal dependencies, it is crucial to simultaneously consider both the global and local representation of nodes from the spatiotemporal perspective. Incorporating these complex dependencies is necessary to improve the performance of GNNs in spatiotemporal applications. Thus, further investigation is required to better model the local and global spatiotemporal intricate dependencies.

\subsubsection{Capturing Spatial Heterogeneity and Temporal Non-Stationarity in graph models}
Spatial heterogeneity and temporal non-stationary widely exist in spatiotemporal data and lead to data distribution shifts. The findings of a benchmark task that employs GNNs to predict air quality in a sensor network indicate that GNN models are more prone to suffering from distribution shift than non-graph-based models \cite{liu2021new}. As a result, it is crucial to pay special attention when utilizing GNNs in practical implementations within the spatiotemporal domain. Domain generalization is a machine learning technique that enable learning a model under varying data distributions. It is an extension of domain adaptation, but with a broad scope \cite{ben2010theory}. While domain generalization has shown remarkable success, little works has been done in the spatiotemporal domain due to difficulty in characterizing spatial heterogeneity and temporal non-stationary. To address the challenge of spatial heterogeneity in spatial data, Yu et al. proposed a spatial interpolation GNN that learns the spatial embedding of each node during the training phase and can infer the spatial embedding of an unseen location during the test phase \cite{yu2022deep}. Bai et al. develop a temporal generalization model to capture the variant relationships in data over time \cite{bai2022temporal}. However, research on spatiotemporal domain generalization using GNNs is still in its early stages, and further investigation is needed.

\subsection{Applications of GNNs in the era of big spatiotemporal data}
\subsubsection{Scalability Challenges in Large Graph Processing}
Training and inference of GNNs at scale remain a challenge using existing deep learning frameworks like TensorFlow, and PyTorch that were designed to handle models with relatively small input samples\cite{jia2020improving}. GNNs operating on a large graph is challenging because the memory complexity is proportional to the total number of nodes and existing frameworks are not optimized for irregular and large input samples \cite{zhang2018gaan}. Recent GNN frameworks such as DGL \cite{wang2019deep} and PyG \cite{fey2019fast} are implemented on top of PyTorch and have the same scalability limitation. NeuGraph \cite{ma2019neugraph} enables multi-GPU training by storing intermediate GNN data in the host CPU DRAM, but it is constrained by the computing resources of a single machine. Due to the lack of system support, the application of GNNs on large-scale graphs has been restricted. However, large graphs are prevalent in the spatiotemporal domain, particularly with the availability of high spatial-temporal resolution data resulting from advanced sensing technologies. Sampling techniques have been introduced to down-sample the graphs to fit on a single device, but this can lead to accuracy loss \cite{jia2020improving}. It remains a challenge to develop a framework that can handle large and complex GNN input in the spatiotemporal domain.
\subsubsection{Applying GNNs to New Spatiotemporal Domains}
The field of GNN in machine learning is relatively new, and its applications on spatiotemporal data are still in their infancy. Recent developments in this area have the potential to make significant contributions to our understanding of information processing in spatiotemporal domains, such as the relationships of regions in brain networks \cite{wein2022forecasting}. While some GNN models have been proposed to handle machine learning tasks on spatiotemporal datasets with an explicit graph structure, such as traffic networks and recommendation systems, limited research exists in the spatiotemporal domain with a latent graph structure, such as climate change studies where teleconnection is common and useful for prediction. In the future, it is essential to investigate how to apply a general spatial-temporal machine learning framework for the graph structure in other practical spatiotemporal applications, such as water consumption and weather prediction. Additionally, it is crucial to explore ways to integrate domain knowledge into the graph construction, aggregation, and prediction processes on both spatial and temporal aspects to make neural networks more interpretable. For instance, by incorporating additional factors like road directions, traffic control, and weather information in a traffic graph model to further enhance performance \cite{zhang2019spatial}, or by embedding the knowledge of physics in a data-driven graph-based model for general spatiotemporal forecasting tasks.

\section{Conclusion}
\noindent This survey provides a comprehensive overview of existing spatiotemporal GNN methods and applications. It offers an extensive analysis of GNN techniques, applications, open challenges and future outlook in the spatiotemporal domain, and summarizes the research from over 150 publications, most of which were published in the last three years. The survey discusses graph construction from spatiotemporal data and proposes a systematic taxonomy of the existing GNN techniques based on the formulated problems and types of methodologies designed for the corresponding spatial and temporal data. The survey also analyzes how these models are combined with other models such as LSTM, GRU, RNN to handle spatial and temporal information simultaneously. Additionally, the survey provides a comprehensive categorization of popular GNN applications, covering domains ranging from natural science to the social sciences. Based on the numerous historical and state-of-the-art works discussed in this survey, the article concludes by highlighting open problems and future trends in this fast-growing field.

\bibliographystyle{ACM-Reference-Format}
\bibliography{sample-base}

\end{document}